\batchmode
\makeatletter
\def\input@path{{/home/fyzhu/link2dropbox/self_Folder/myWorksOnDropboxs/201702_SigKDD_CohesionDiscovery4onlineGraphRL_self/IEEE_TPAMI//}}
\makeatother
\documentclass[10pt,journal,compsoc]{IEEEtran}
\usepackage[latin9]{inputenc}
\usepackage{color}
\definecolor{page_backgroundcolor}{rgb}{1, 1, 1}
\pagecolor{page_backgroundcolor}
\usepackage{array}
\usepackage{float}
\usepackage{bm}
\usepackage{multirow}
\usepackage{amsmath}
\usepackage{amsthm}
\usepackage{graphicx}
\usepackage[unicode=true,pdfusetitle,
 bookmarks=true,bookmarksnumbered=false,bookmarksopen=false,
 breaklinks=true,pdfborder={0 0 0},backref=false,colorlinks=true]
 {hyperref}

\makeatletter

\providecommand{\tabularnewline}{\\}
\floatstyle{ruled}
\newfloat{algorithm}{tbp}{loa}
\providecommand{\algorithmname}{Algorithm}
\floatname{algorithm}{\protect\algorithmname}

\theoremstyle{plain}
\newtheorem{thm}{\protect\theoremname}
\theoremstyle{plain}
\newtheorem{lem}[thm]{\protect\lemmaname}

\ifCLASSOPTIONcompsoc
  \usepackage[nocompress]{cite}
\else
  \usepackage{cite}
\fi
\ifCLASSINFOpdf
\else
\fi
\usepackage{algorithm}
\usepackage{algorithmic}
\usepackage{amssymb}
\usepackage[square,sort&compress,numbers]{natbib}

\makeatother

\providecommand{\lemmaname}{Lemma}
\providecommand{\theoremname}{Theorem}

\begin{document}
\global\long\def\mtbfA{\mathbf{A}}
 \global\long\def\mtbfa{\mathbf{a}}
 \global\long\def\mebfA{\bar{\mtbfA}}
 \global\long\def\mebfa{\bar{\mtbfa}}

\global\long\def\mhbfA{\widehat{\mathbf{A}}}
 \global\long\def\mhbfa{\widehat{\mathbf{a}}}
 \global\long\def\mtcalA{\mathcal{A}}
 \global\long\def\mtbbA{\mathbb{A}}

\global\long\def\mtbfB{\mathbf{B}}
 \global\long\def\mtbfb{\mathbf{b}}
 \global\long\def\mebfB{\bar{\mtbfB}}
 \global\long\def\mebfb{\bar{\mtbfb}}

\global\long\def\mhbfB{\widehat{\mathbf{B}}}
 \global\long\def\mhbfb{\widehat{\mathbf{b}}}
 \global\long\def\mtcalB{\mathcal{B}}
 \global\long\def\mtbbB{\mathbb{B}}

\global\long\def\mtbfC{\mathbf{C}}
 \global\long\def\mtbfc{\mathbf{c}}
 \global\long\def\mebfC{\bar{\mtbfC}}
 \global\long\def\mebfc{\bar{\mtbfc}}

\global\long\def\mhbfC{\widehat{\mathbf{C}}}
 \global\long\def\mhbfc{\widehat{\mathbf{c}}}
 \global\long\def\mtcalC{\mathcal{C}}
 \global\long\def\mtbbC{\mathbb{C}}

\global\long\def\mtbfD{\mathbf{D}}
 \global\long\def\mtbfd{\mathbf{d}}
 \global\long\def\mebfD{\bar{\mtbfD}}
 \global\long\def\mebfd{\bar{\mtbfd}}

\global\long\def\mhbfD{\widehat{\mathbf{D}}}
 \global\long\def\mhbfd{\widehat{\mathbf{d}}}
 \global\long\def\mtcalD{\mathcal{D}}
 \global\long\def\mtbbD{\mathbb{D}}

\global\long\def\mtbfE{\mathbf{E}}
 \global\long\def\mtbfe{\mathbf{e}}
 \global\long\def\mebfE{\bar{\mtbfE}}
 \global\long\def\mebfe{\bar{\mtbfe}}

\global\long\def\mhbfE{\widehat{\mathbf{E}}}
 \global\long\def\mhbfe{\widehat{\mathbf{e}}}
 \global\long\def\mtcalE{\mathcal{E}}
 \global\long\def\mtbbE{\mathbb{E}}

\global\long\def\mtbfF{\mathbf{F}}
 \global\long\def\mtbff{\mathbf{f}}
 \global\long\def\mebfF{\bar{\mathbf{F}}}
 \global\long\def\mebff{\bar{\mathbf{f}}}

\global\long\def\mhbfF{\widehat{\mathbf{F}}}
 \global\long\def\mhbff{\widehat{\mathbf{f}}}
 \global\long\def\mtcalF{\mathcal{F}}
 \global\long\def\mtbbF{\mathbb{F}}

\global\long\def\mtbfG{\mathbf{G}}
 \global\long\def\mtbfg{\mathbf{g}}
 \global\long\def\mebfG{\bar{\mathbf{G}}}
 \global\long\def\mebfg{\bar{\mathbf{g}}}

\global\long\def\mhbfG{\widehat{\mathbf{G}}}
 \global\long\def\mhbfg{\widehat{\mathbf{g}}}
 \global\long\def\mtcalG{\mathcal{G}}
 \global\long\def\mtbbG{\mathbb{G}}

\global\long\def\mtbfH{\mathbf{H}}
 \global\long\def\mtbfh{\mathbf{h}}
 \global\long\def\mebfH{\bar{\mathbf{H}}}
 \global\long\def\mebfh{\bar{\mathbf{h}}}

\global\long\def\mhbfH{\widehat{\mathbf{H}}}
 \global\long\def\mhbfh{\widehat{\mathbf{h}}}
 \global\long\def\mtcalH{\mathcal{H}}
 \global\long\def\mtbbH{\mathbb{H}}

\global\long\def\mtbfI{\mathbf{I}}
 \global\long\def\mtbfi{\mathbf{i}}
 \global\long\def\mebfI{\bar{\mathbf{I}}}
 \global\long\def\mebfi{\bar{\mathbf{i}}}

\global\long\def\mhbfI{\widehat{\mathbf{I}}}
 \global\long\def\mhbfi{\widehat{\mathbf{i}}}
 \global\long\def\mtcalI{\mathcal{I}}
 \global\long\def\mtbbI{\mathbb{I}}

\global\long\def\mtbfJ{\mathbf{J}}
 \global\long\def\mtbfj{\mathbf{j}}
 \global\long\def\mebfJ{\bar{\mathbf{J}}}
 \global\long\def\mebfj{\bar{\mathbf{j}}}

\global\long\def\mhbfJ{\widehat{\mathbf{J}}}
 \global\long\def\mhbfj{\widehat{\mathbf{j}}}
 \global\long\def\mtcalJ{\mathcal{J}}
 \global\long\def\mtbbJ{\mathbb{J}}

\global\long\def\mtbfK{\mathbf{K}}
 \global\long\def\mtbfk{\mathbf{k}}
 \global\long\def\mebfK{\bar{\mathbf{K}}}
 \global\long\def\mebfk{\bar{\mathbf{k}}}

\global\long\def\mhbfK{\widehat{\mathbf{K}}}
 \global\long\def\mhbfk{\widehat{\mathbf{k}}}
 \global\long\def\mtcalK{\mathcal{K}}
 \global\long\def\mtbbK{\mathbb{K}}

\global\long\def\mtbfL{\mathbf{L}}
 \global\long\def\mtbfl{\mathbf{l}}
 \global\long\def\mebfL{\bar{\mathbf{L}}}
 \global\long\def\mebfl{\bar{\mathbf{l}}}

\global\long\def\mhbfL{\widehat{\mathbf{K}}}
 \global\long\def\mhbfl{\widehat{\mathbf{k}}}
 \global\long\def\mtcalL{\mathcal{L}}
 \global\long\def\mtbbL{\mathbb{L}}

\global\long\def\mtbfM{\mathbf{M}}
 \global\long\def\mtbfm{\mathbf{m}}
 \global\long\def\mebfM{\bar{\mathbf{M}}}
 \global\long\def\mebfm{\bar{\mathbf{m}}}

\global\long\def\mhbfM{\widehat{\mathbf{M}}}
 \global\long\def\mhbfm{\widehat{\mathbf{m}}}
 \global\long\def\mtcalM{\mathcal{M}}
 \global\long\def\mtbbM{\mathbb{M}}

\global\long\def\mtbfN{\mathbf{N}}
 \global\long\def\mtbfn{\mathbf{n}}
 \global\long\def\mebfN{\bar{\mathbf{N}}}
 \global\long\def\mebfn{\bar{\mathbf{n}}}

\global\long\def\mhbfN{\widehat{\mathbf{N}}}
 \global\long\def\mhbfn{\widehat{\mathbf{n}}}
 \global\long\def\mtcalN{\mathcal{N}}
 \global\long\def\mtbbN{\mathbb{N}}

\global\long\def\mtbfO{\mathbf{O}}
 \global\long\def\mtbfo{\mathbf{o}}
 \global\long\def\mebfO{\bar{\mathbf{O}}}
 \global\long\def\mebfo{\bar{\mathbf{o}}}

\global\long\def\mhbfO{\widehat{\mathbf{O}}}
 \global\long\def\mhbfo{\widehat{\mathbf{o}}}
 \global\long\def\mtcalO{\mathcal{O}}
 \global\long\def\mtbbO{\mathbb{O}}

\global\long\def\mtbfP{\mathbf{P}}
 \global\long\def\mtbfp{\mathbf{p}}
 \global\long\def\mebfP{\bar{\mathbf{P}}}
 \global\long\def\mebfp{\bar{\mathbf{p}}}

\global\long\def\mhbfP{\widehat{\mathbf{P}}}
 \global\long\def\mhbfp{\widehat{\mathbf{p}}}
 \global\long\def\mtcalP{\mathcal{P}}
 \global\long\def\mtbbP{\mathbb{P}}

\global\long\def\mtbfQ{\mathbf{Q}}
 \global\long\def\mtbfq{\mathbf{q}}
 \global\long\def\mebfQ{\bar{\mathbf{Q}}}
 \global\long\def\mebfq{\bar{\mathbf{q}}}

\global\long\def\mhbfQ{\widehat{\mathbf{Q}}}
 \global\long\def\mhbfq{\widehat{\mathbf{q}}}
\global\long\def\mtcalQ{\mathcal{Q}}
 \global\long\def\mtbbQ{\mathbb{Q}}

\global\long\def\mtbfR{\mathbf{R}}
 \global\long\def\mtbfr{\mathbf{r}}
 \global\long\def\mebfR{\bar{\mathbf{R}}}
 \global\long\def\mebfr{\bar{\mathbf{r}}}

\global\long\def\mhbfR{\widehat{\mathbf{R}}}
 \global\long\def\mhbfr{\widehat{\mathbf{r}}}
\global\long\def\mtcalR{\mathcal{R}}
 \global\long\def\mtbbR{\mathbb{R}}

\global\long\def\mtbfS{\mathbf{S}}
 \global\long\def\mtbfs{\mathbf{s}}
 \global\long\def\mebfS{\bar{\mathbf{S}}}
 \global\long\def\mebfs{\bar{\mathbf{s}}}

\global\long\def\mhbfS{\widehat{\mathbf{S}}}
 \global\long\def\mhbfs{\widehat{\mathbf{s}}}
\global\long\def\mtcalS{\mathcal{S}}
 \global\long\def\mtbbS{\mathbb{S}}

\global\long\def\mtbfT{\mathbf{T}}
 \global\long\def\mtbft{\mathbf{t}}
 \global\long\def\mebfT{\bar{\mathbf{T}}}
 \global\long\def\mebft{\bar{\mathbf{t}}}

\global\long\def\mhbfT{\widehat{\mathbf{T}}}
 \global\long\def\mhbft{\widehat{\mathbf{t}}}
 \global\long\def\mtcalT{\mathcal{T}}
 \global\long\def\mtbbT{\mathbb{T}}

\global\long\def\mtbfU{\mathbf{U}}
 \global\long\def\mtbfu{\mathbf{u}}
 \global\long\def\mebfU{\bar{\mathbf{U}}}
 \global\long\def\mebfu{\bar{\mathbf{u}}}

\global\long\def\mhbfU{\widehat{\mathbf{U}}}
 \global\long\def\mhbfu{\widehat{\mathbf{u}}}
 \global\long\def\mtcalU{\mathcal{U}}
 \global\long\def\mtbbU{\mathbb{U}}

\global\long\def\mtbfV{\mathbf{V}}
 \global\long\def\mtbfv{\mathbf{v}}
 \global\long\def\mebfV{\bar{\mathbf{V}}}
 \global\long\def\mebfv{\bar{\mathbf{v}}}

\global\long\def\mhbfV{\widehat{\mathbf{V}}}
 \global\long\def\mhbfv{\widehat{\mathbf{v}}}
\global\long\def\mtcalV{\mathcal{V}}
 \global\long\def\mtbbV{\mathbb{V}}

\global\long\def\mtbfW{\mathbf{W}}
 \global\long\def\mtbfw{\mathbf{w}}
 \global\long\def\mebfW{\bar{\mathbf{W}}}
 \global\long\def\mebfw{\bar{\mathbf{w}}}

\global\long\def\mhbfW{\widehat{\mathbf{W}}}
 \global\long\def\mhbfw{\widehat{\mathbf{w}}}
 \global\long\def\mtcalW{\mathcal{W}}
 \global\long\def\mtbbW{\mathbb{W}}

\global\long\def\mtbfX{\mathbf{X}}
 \global\long\def\mtbfx{\mathbf{x}}
 \global\long\def\mebfX{\bar{\mtbfX}}
 \global\long\def\mebfx{\bar{\mtbfx}}

\global\long\def\mhbfX{\widehat{\mathbf{X}}}
 \global\long\def\mhbfx{\widehat{\mathbf{x}}}
 \global\long\def\mtcalX{\mathcal{X}}
 \global\long\def\mtbbX{\mathbb{X}}

\global\long\def\mtbfY{\mathbf{Y}}
 \global\long\def\mtbfy{\mathbf{y}}
\global\long\def\mebfY{\bar{\mathbf{Y}}}
 \global\long\def\mebfy{\bar{\mathbf{y}}}

\global\long\def\mhbfY{\widehat{\mathbf{Y}}}
 \global\long\def\mhbfy{\widehat{\mathbf{y}}}
 \global\long\def\mtcalY{\mathcal{Y}}
 \global\long\def\mtbbY{\mathbb{Y}}

\global\long\def\mtbfZ{\mathbf{Z}}
 \global\long\def\mtbfz{\mathbf{z}}
 \global\long\def\mebfZ{\bar{\mathbf{Z}}}
 \global\long\def\mebfz{\bar{\mathbf{z}}}

\global\long\def\mhbfZ{\widehat{\mathbf{Z}}}
 \global\long\def\mhbfz{\widehat{\mathbf{z}}}
\global\long\def\mtcalZ{\mathcal{Z}}
 \global\long\def\mtbbZ{\mathbb{Z}}

\global\long\def\mtth{\text{th}}

\global\long\def\mtbfzero{\mathbf{0}}
 \global\long\def\mtbfone{\mathbf{1}}

\global\long\def\mttrace{\text{Tr}}

\global\long\def\mttotalVariation{\text{TV}}

\global\long\def\mtexpect{\mathbb{E}}

\global\long\def\mtdet{\text{det}}

\global\long\def\mtvec{\mathbf{\text{vec}}}

\global\long\def\mtvar{\mathbf{\text{var}}}

\global\long\def\mtcov{\mathbf{\text{cov}}}

\global\long\def\mtsubTo{\mathbf{\text{s.t.}}}

\global\long\def\mtfor{\text{for}}

\global\long\def\mtrank{\text{rank}}

\global\long\def\mtrankn{\text{rankn}}

\global\long\def\mtdiag{\mathbf{\text{diag}}}

\global\long\def\mtsign{\mathbf{\text{sign}}}

\global\long\def\mtloss{\mathbf{\text{loss}}}

\global\long\def\mtwhen{\text{when}}

\global\long\def\mtwhere{\text{where}}

\global\long\def\mtif{\text{if}}

\title{Cohesion-based Online Actor-Critic Reinforcement Learning for mHealth
Intervention}

\author{Feiyun Zhu$^{\star,\ddagger}$, Peng Liao$^{\ddagger}$, Xinliang
Zhu$^{\star}$, Yaowen Yao$^{\star}$, Junzhou Huang$^{\star}$ \thanks{$\star$ Feiyun~Zhu, Xinliang Zhu, Yaowen Yao and Junzhou Huang are
with the Department of CSE in the University of Texas at Arlington.
$\ddagger$ Feiyun Zhu and Peng Liao are with the Department of Statistics
in the University of Michigan.}}
\maketitle
\begin{abstract}
In the wake of the vast population of smart device users worldwide,
mobile health (mHealth) technologies are hopeful to generate positive
and wide influence on people's health. They are able to provide flexible,
affordable and portable health guides to device users. Current online
decision-making methods for mHealth assume that the users are completely
heterogeneous. They share no information among users and learn a separate
policy for each user. However, data for each user is very limited
in size to support the separate online learning, leading to unstable
policies that contain lots of variances. Besides, we find the truth
that a user may be similar with some, but not all, users, and connected
users tend to have similar behaviors. In this paper, we propose a
network cohesion constrained (actor-critic) Reinforcement Learning
(RL) method for mHealth. The goal is to explore how to share information
among similar users to better convert the limited user information
into sharper learned policies. To the best of our knowledge, this
is the first online actor-critic RL for mHealth and first network
cohesion constrained (actor-critic) RL method in all applications.
The network cohesion is important to derive effective policies. We
come up with a novel method to learn the network by using the warm
start trajectory, which directly reflects the users' property. The
optimization of our model is difficult and very different from the
general supervised learning due to the indirect observation of values.
As a contribution, we propose two algorithms for the proposed online
RLs. Apart from mHealth, the proposed methods can be easily applied
or adapted to other health-related tasks. Extensive experiment results
on the HeartSteps dataset demonstrates that in a variety of parameter
settings, the proposed two methods obtain obvious improvements over
the state-of-the-art methods. \end{abstract}


\section{Introduction}

With billions of smart device\footnote{i.e. smartphones and wearable devices, such as Fitbit Fuelband and
Jawbone etc.} users globally, it is increasingly popular among the scientist community
to make use of the state-of-the-art articial intelligence and mobile
health technologies to leverage supercomputers and big data to facilicate
the prediction of healthcare tasks\ \cite{huitian_2014_NIPS_ActCriticBandit4JITAI,huitian_2016_PhdThesis_actCriticAlgorithm,SusanMurphy_2016_CORR_BatchOffPolicyAvgRwd,PengLiao_2015_Proposal_offPolicyRL,yaoyao_2017_MICCAI,xinliang_2017_CVPR_WSISA,zhengxu_2017_ACMBCB}.
In this paper, the goal of mobile health (mHealth) is to make use
of various smart devices as great platforms to collect and analyze
raw data (weather, location, social activity, stress, etc.). Based
on that, the aim is to provide effective intervention that helps users
to change to or adapt to healthy behaviors, such as reducing the alcohol
abuse\ \cite{Gustafson_2014_JAMA_drinking,Witkiewitz_2014_JAB_drinkingSmoking}
and promoting physical activities\ \cite{Abby_2013_PlosONE_mobileIntervention}.
The traditional adaptive treatment has restrictions on the time, location
and frequency---patients have to visit the doctor's office for treatments.
Compared with them, mHealth is more affordable, portable and much
more flexible in the sense that smart devices allow for the real-time
collection and analysis of data as well as in-time delivery of interventions.
Thus, mHealth technologies are widely used in lots of health-related
tasks, such as physical activity\ \cite{Abby_2013_PlosONE_mobileIntervention},
eating disorders\ \cite{Bauer_2012_JCCP_eatingDisorder}, alcohol
use\ \cite{Gustafson_2014_JAMA_drinking,Witkiewitz_2014_JAB_drinkingSmoking},
mental illness\cite{Depp_2010_JNMD_mentalIllness,Ben_2013_APMHMHSRZZZZ_mentalIllness},
obesity/weight management \cite{Patrick_2009_JMIR_weightManagement}.

Formally, the mHealth intervention is modeled as a sequential decision
making (SDM) problem. It aims to learn the optimal policy to determine
when, where and how to deliver the intervention\ \cite{huitian_2014_NIPS_ActCriticBandit4JITAI,PengLiao_2015_Proposal_offPolicyRL,SusanMurphy_2016_CORR_BatchOffPolicyAvgRwd}
to best serve users. This is a new research topic that lacks of methodological
guidance. In 2014, Lei\ \cite{huitian_2014_NIPS_ActCriticBandit4JITAI}
made a first attempt to formulate the mHealth intervention as an online
actor-critic contextual bandit problem. Lei's method served a good
starting point for the mHealth study. However, this method did not
consider the important delayed effect in the SDM---the current action
may influence not only the immediate reward but also the next states
and, through that, all the subsequent rewards\ \cite{Sutton_2012_Book_ReinforcementLearning,YihongLi_2010_WWW_contextualBandit4newsArticleRecommend}.
Dr. Murphy\ \cite{SusanMurphy_2016_CORR_BatchOffPolicyAvgRwd} proposed
an average reward based RL to consider the delayed effect in the mHealth.
However, those two methods rely on some ideal assumptions. They either
assume that all the users are completely homogenous or completely
heterogeneous. We find the truth lying between those extremes: a user
might be similar with some, but not all, users. Their methods are
easy to bring in too much bias or too much variance. Besides, \cite{SusanMurphy_2016_CORR_BatchOffPolicyAvgRwd}
is in the batch learning setting, which is different from this paper's
focuses.

Recently, Dr. Cesa-Bianchi\ \cite{Bianchi_2013_NIPS_GangOfBandit}
proposed a contextual bandit algorithm that considers the network
information. It is for the recommendation system, which is very different
from the mHealth task. Besides, there are three drawbacks making the
method in\ \cite{Bianchi_2013_NIPS_GangOfBandit} impractical for
the mHealth: (1) Cesa-Bianchi's method focues on the bandit algorithm.
It doesn't consider the important delayed effect in mHealth. (2) They
assume the network information is given beforehand from the social
information. The given network may not be targeted for the mHealth
study. There is lots of misleading network information for the mHealth
study\ \cite{Bianchi_2013_NIPS_GangOfBandit,Claudio_2014_ICML_olineClusteringBandits,Alexandra_2016_AISTATS_GraphBandits}.
(3) In their work, however, it is unable to control the amount of
information shared among linked users, which is not flexible for the
mHealth study\ \cite{fyzhu_2014_IJPRS_SSNMF,haichangLi_2016_IJRS_LablePropagationHyperClassification}.

In this paper, we propose a cohesion-based reinforcement learning
for the mHealth and derive two algorithms. It is in an online, actor-critic
setting. The aim is to explore how to share information across similar
users in order to improve the performance. The main contributions
of this paper are summarized as follows: (\textbf{1}) to the best
of our knowledge, this is the first online (actor-critic) RL method
for the mHealth. (\textbf{2}) Current evidence verifies the wide existence
of networks among users\ \cite{Tianxi_2016_CORR_PredictModels4NetworkLinkedData,fyzhu_2014_IJPRS_SSNMF,haichangLi_2016_IJRS_LablePropagationHyperClassification}.
We improve the online RL by considering the network cohesion among
users. Such improvement makes it the first network constrained (actor-critic)
RL method to the best of our knowledge. It is able to relieve the
tough problem of current online decision-making methods for the mHealth
by reducing variance at the cost of inducing bias. Current online
RL learns a separate policy for each user. However, there are too
few of samples to support the separate online learning, which leads
to unsatisfactory interventions (policies) for the users. (\textbf{3})
Our method doesn't require the given network cohesion. We propose
a method to learn the network intentionally for the mHealth study.
It makes use of the warm start trajectories in the online learning,
which are expected to represent the users' properties. (\textbf{4})
Compared with\ \cite{Bianchi_2013_NIPS_GangOfBandit}, the proposed
method has a tuning parameter, which allows us to control how much
information we should share with similar users. It is worth mentioning
that our method may not be limited to mHealth. It can be applied to
other health-related tasks. Extensive experiment results on the HeartSteps
dataset verifies that our method can achieve clear improvement over
the Separate-RL.

\section{Preliminaries}

\subsection{Markov Decision Process (MDP)}

We assume the mHealth intervention is a Markov Decision Process (MDP)\ \cite{Geist_2013_TNNLS_RL_valueFunctionApproximation,Grondman_2012_IEEEts_surveyOfActorCriticRL,Pednault_2002_SIGKDD_CostSensitiveRL,Michail_2003_JMLR_LSPI_LSTDQ}
that consists of a 5-tuple $\left\{ \mtcalS,\mtcalA,\mtcalP,\mtcalR,\gamma\right\} $,
where $\mtcalS$ is the state space and $\mtcalA$ is the action space.
$\mtcalP\!\!:\!\mtcalS\!\times\!\mtcalA\!\times\!\mtcalS\!\mapsto\!\left[0,1\right]$
is the state transition model in which $\mtcalP\left(s,a,s'\right)$
indicates the probability of transiting from one state $s$ to another
$s'$ after taking action $a$; $\mtcalR\left(s,a,s'\right)$ is the
corresponding immediate reward for such transition where $\mtcalR\!:\mtcalS\times\mtcalA\times\mtcalS\mapsto\mtbbR$.
For simplicity, the expected immediate reward $\mtcalR\left(s,a\right)=\mtexpect_{s'\sim\mtcalP}\left[\mtcalR\left(s,a,s'\right)\right]$
is assumed to be bounded over the state and action spaces. $\gamma\in[0,1)$
is the discount factor that reduces the influence of future rewards.
To allow for the matrix operators, the state space $\mtcalS$ and
action space $\mtcalA$ are assumed to be finite, though very large
in mHealth.

The policy of an MDP is to choose actions for any state $s\in\mtcalS$
in the system\ \cite{Grondman_2012_IEEEts_surveyOfActorCriticRL,Sutton_2012_Book_ReinforcementLearning}.
There are two types of policies: (1) the deterministic policy $\pi:\mtcalS\mapsto\mtcalA$
selects an action directly for the state, and (2) the stochastic policy
$\pi:s\in\mtcalS\mapsto\pi\left(\cdot\mid s\right)\in\mtcalP\left(\mtcalA\right)$
chooses the action for any state $s$ by providing $s$ with a probability
distribution over all the possible actions\ \cite{Geist_2013_TNNLS_RL_valueFunctionApproximation}.
In mHealth, the stochastic policy is preferred due to two reasons:
(a) current evidence shows that some randomness in the action is likely
to draw users' interest, thus helpful to reduce the intervention burden/habituation\ \cite{Epstein_2009_AJCN_varietyInfluencesHab,huitian_2014_NIPS_ActCriticBandit4JITAI,SusanMurphy_2016_CORR_BatchOffPolicyAvgRwd};
(b) though some deterministic policy is theoretically optimal for
the MDP, however, we do not know where it is for the large state space
on the one hand and the MDP is a simplification for the complex behavioral
process on the other; some variation may be helpful to explore the
system and search for a desirable policy\ \cite{SusanMurphy_2016_CORR_BatchOffPolicyAvgRwd}.
We consider the parameterized stochastic policy, $\pi_{\theta}\left(a\mid s\right)$,
where $\theta\in\mtbbR^{m}$ is the unknown parameter. Such policy
is interpretable in the sense that we could know the key features
that contribute most to the policy by analyzing the estimated $\widehat{\theta}$,
which is important to behavior scientists for the state (feature)
design\ \cite{huitian_2014_NIPS_ActCriticBandit4JITAI,SusanMurphy_2016_CORR_BatchOffPolicyAvgRwd}.

In RL, value is a core concept that quantifies the quality of a policy
$\pi$\ \cite{Sutton_2012_Book_ReinforcementLearning}. There are
two definitions of values: the state value and the state-action ($Q$-)
value\ \cite{Abe_2004_SIGKDD_CrossChannelRL}. In mHealth, the $Q$-value
is considered because the model (i.e. state transition and immediate
reward) is assumed to be unknown, and $Q$-value allows for action
selection without knowing the model while the state value requires
the model for the action selection\ \cite{Michail_2003_JMLR_LSPI_LSTDQ}.
Formally, the $Q$-value $Q^{\pi}\left(s,a\right)\in\mtbbR^{\left|\mtcalS\right|\times\left|\mtcalA\right|}$
measures the total amount of rewards an agent can obtain when starting
from state $s$, first choosing action $a$ and then following the
policy $\pi$. Specially, the discounted reward is one of the most
commonly used value measures 
\begin{equation}
Q^{\pi}\left(s,a\right)=\mtexpect_{a_{i}\sim\pi,s_{i}\sim\mtcalP}\left\{ \sum_{i=0}^{\infty}\gamma^{i}r_{i}\mid s_{0}=s,a_{0}=a\right\} .\label{eq:Q_value}
\end{equation}

The goal of RL is to learn an optimal policy $\pi^{*}$ that maximizes
the $Q$-value for all the state-action pairs via interactions with
the dynamic system\ \cite{Geist_2013_TNNLS_RL_valueFunctionApproximation}.
The objective is ${\displaystyle \theta^{*}=\arg\max_{\theta}\widehat{J}\left(\theta\right)},$
where 
\begin{equation}
\widehat{J}\left(\theta\right)=\sum_{s\in\mtcalS}d_{\text{ref}}\left(s\right)\sum_{a\in\mtcalA}\pi_{\theta}\left(a\mid s\right)Q^{\pi_{\theta}}\left(s,a\right)\label{eq:actor-objective_thoery}
\end{equation}
and $d_{\text{ref}}\left(s\right)$ is the reference distribution
of states (e.g. the distribution of initial states); $Q^{\pi_{\theta}}$
is the value for the policy $\pi_{\theta}$. According to\ \eqref{eq:actor-objective_thoery},
we have to learn the $Q^{\pi_{\theta}}$ for all the state-action
pairs to determine the objective\ \eqref{eq:actor-objective_thoery}
and, after then, to improve the policy. Thus in this paper, we employ
the actor-critic algorithm. It is an alternating updating algorithm
between two steps untill convergence. At each iteration, the critic
updating estimates the $Q$-value function (i.e. policy evaluation,
cf. Section\ \ref{sub:BellmanEquation_MC_TD} and\ \ref{sub:CriticUpdating_LSTDQ})
for the lastest policy; the actor updating (i.e. policy improvement,
cf. Section\ \ref{sub:ActorUpdating_fmincon}) learns a better policy
based on the newly estimated $Q$-value. Moreover, the actor-critic
algorithm has great properties of quick convergence with low variance
and learning continuous policies\ \cite{Grondman_2012_IEEEts_surveyOfActorCriticRL}.

\subsection{Bellman Equation and Q-value Estimation}

\label{sub:BellmanEquation_MC_TD} It is well known that due to the
Markovian property, the $Q$-value satisfies the linear Bellman equation\ \cite{Abe_2010_SIGKDD_optimizingDebt}
for any policy $\pi$:
\begin{align*}
Q^{\pi}\left(s,a\right)=\  & \mtcalR\left(s,a\right)+\gamma\sum_{s^{'}\in\mtcalS}\mtcalP\left(s,a,s'\right)\sum_{a^{'}\in\mtcalA}\pi\left(a'\mid s'\right)Q^{\pi}\left(s',a'\right).
\end{align*}
It has the matrix form as
\begin{equation}
\mtbfq^{\pi}=\mtbfr+\gamma\mtbfP\Pi_{\pi}\mtbfq^{\pi},\label{eq:Q_linearBellmanEquation_detail}
\end{equation}
where $\mtbfq^{\pi}$ and $\mtbfr$ are vectors both with $\left|\mtcalS\right|\left|\mtcalA\right|$
elements; $\mtbfP\in\mtbbR^{\left|\mtcalS\right|\left|\mtcalA\right|\times\left|\mtcalS\right|}$
is the stochastic state transition matrix, in which $P\left(\left(s,a\right),s'\right)=\mtcalP\left(s,a,s'\right)$;
$\Pi_{\pi}\in\mtbbR^{\left|\mtcalS\right|\times\left|\mtcalS\right|\left|\mtcalA\right|}$
is the stochastic policy matrix, where $\Pi_{\pi}\left(s,\left(s,a\right)\right)=\pi\left(a\mid s\right)$\ \cite{Michail_2003_JMLR_LSPI_LSTDQ}.
Once both the reward and the state transition models are given\ \cite{AndrewNg_2009_ICML_RLsparity},
it is easy to obtain the analytical solution as $\mtbfq^{\pi}=\left(\mtbfI-\gamma\mtbfP\Pi_{\pi}\right)^{-1}\mtbfr$.

However, there are two factors making it impossible to have the analytical
solution for the $Q$-value estimation: (a) in mHealth, both reward
$\mtbfr$ and state transition $\mtcalP$ (i.e. $\mtbfP$) models
are unknown. (b) the state space in mHealth is usually very large
or even infinite, which makes it impossible to directly learn the
$Q$-value due to lack of observations for sharper learning and too
high storage requirements, i.e. $O\left(\left|\mtcalS\right|\left|\mtcalA\right|\right)$
to only store the $Q$-value table. We resolve these problems via
the parameterized function approximation, which assumes that $Q^{\pi}$
is in a low dimensional space: $Q^{\pi}\approx Q_{\mtbfw}=\mtbfw^{\intercal}\mtbfx\left(s,a\right)$,
where $\mtbfw\in\mtbbR^{u}$ is the unknown variable and $\mtbfx\left(s,a\right)$
is a feature processing step that combines information in the state
and action. We then learn the value $Q_{\mtbfw}$ from observations
via a supervised learning paramdigm, which, however, is much more
challenging than the general supervised learning since the $Q$-value
is not directly observed\ \cite{Geist_2013_TNNLS_RL_valueFunctionApproximation}.
As a direct solution, the Monte Carlo (MC) method draws very deep
trajectories to obtain the observation of actual $Q$ value. Although
MC can provide an unbiased estimation of $Q_{\mtbfw}$, it is not
suitable for mHealth since MC can't learn from the incomplete trajectory\ \cite{Sutton_2012_Book_ReinforcementLearning}.
Such case requires massive sampling from users, which, however, is
very labor-intensive and expensive in time. As a central idea of RL\ \cite{Sutton_2012_Book_ReinforcementLearning},
the temporal-difference (TD) learning is able to make use of the Bellman
equation\ \eqref{eq:Q_linearBellmanEquation_detail} and to learn
the value from the incomplete trajectories. The learned result of
TD has the property of low variance.

\subsection{The critic updating: Least-Squares TD for $Q$-value (LSTD$Q$) Estimation}

\label{sub:CriticUpdating_LSTDQ}

In mHealth, though the data for all users is abundant, the data for
each user is limited in size. We employ the least-square TD for the
Q-value (LSTD$Q$) estimation, due to its advantage of efficient use
of samples over the pure temporal-difference algorithms\ \cite{Michail_2003_JMLR_LSPI_LSTDQ,Sakuma_2008_ICML_PrivacyPreservingRL}.
The goal of LSTD$Q$ is to learn a $Q_{\mtbfw}$ to approximately
satisfy the Bellman equation\ \eqref{eq:Q_linearBellmanEquation_detail},
by minimizing the TD error\ \cite{AndrewNg_2009_ICML_RLsparity}
as 
\begin{equation}
\mtbfw=\mhbfh=\min_{\mtbfh\in\mtbbR^{K}}\left\Vert \mebfX^{\intercal}\mtbfh-\left(\mebfr+\gamma\mtbfP\Pi_{\pi}\mebfX^{\intercal}\mtbfw\right)\right\Vert _{D}^{2},\label{eq:LSTDQ_distribution}
\end{equation}
where $\mtbfw=\mhbfh$ is a fixed point problem and $\mhbfh$ is a
function of $\mtbfw$; $\mebfX$ is a designed matrix consisting of
all the state and action pairs in the MDP; $D$ describes the distributions
over the state and action pairs.

Since the state transition $\mtbfP$ is unknown and $\mebfX$ is too
large to form in mHealth, we can not directly solve\ \eqref{eq:LSTDQ_distribution}.
Instead, we have to make use of the trajectories collected from $N$
users, i.e. $\mtcalD=\left\{ \mtcalD_{n}\right\} _{n=1}^{N}$, where
$\mtcalD_{n}=\left\{ \mtcalU_{i}=\left(s_{i},a_{i,}r_{i},s_{i}'\right)\mid i=0,\cdots,t\right\} $
summarizes all the $t+1$ tuples for the $n$-th user and $\mtcalU_{i}$
is the $i$-th tuple in $\mtcalD_{n}$.

Current online contextual bandit (i.e. a special RL with $\gamma=0$)
methods for mHealth assume that all users are completely heterogeneous.
They share no information and run a separate algorithm for each user\ \cite{huitian_2014_NIPS_ActCriticBandit4JITAI}.
Following this idea, we extend\ \cite{huitian_2014_NIPS_ActCriticBandit4JITAI}
to the separate RL setting. The objective for the $n$-th user is
defined as 
\begin{align}
\mtbfw_{n} & =\mhbfh_{n}=\arg\min_{\mtbfh_{n}}\sum_{\mtcalU_{i}\in\mtcalD_{n}}\left\Vert \mtbfx{}_{i}^{\intercal}\mtbfh_{n}-\left(r_{i}+\gamma\mtbfy{}_{i}^{\intercal}\mtbfw_{n}\right)\right\Vert _{2}^{2}
\end{align}
$\mtfor\ n\in\left\{ 1,\cdots,N\right\} $, where $\mtbfx_{i}=\mtbfx\left(s_{i},a_{i}\right)$
is the value feature at time $i$ and $\mtbfy{}_{i}=\sum_{a'\in\mtcalA}\mathbf{x}\left(s_{i}',a'\right)\pi_{\theta}\left(a'\mid s_{i}'\right)$
is the value feature at the next time point. For the sake of easy
derivation, we define the following matrices to store the actual observations
\begin{align}
\mtbfX_{n} & =\left[\mathbf{x}\left(s_{1},a_{1}\right),\mathbf{x}\left(s_{2},a_{2}\right)\cdots,\mathbf{x}\left(s_{t},a_{t}\right)\right]\in\mathbb{R}^{u\times t}\nonumber \\
\mtbfY_{n} & =\left[\mtbfy\left(s_{1}';\theta_{n}\right),\cdots,\mtbfy\left(s_{t}';\theta_{n}\right)\right]\in\mathbb{R}^{u\times t}\label{eq:featureConstruction_4_value}\\
\mtbfr_{n} & =\left[r_{1},r_{2},\cdots,r_{t}\right]^{\intercal}\in\mathbb{R}^{t},\nonumber 
\end{align}
where $u$ is the length of the sample feature for the $Q$-value
approximation, $t$ is the current time point in the online RL learning
procedure (i.e. the current trajectory length), $\mtbfy\left(s_{i}';\theta_{n}\right)=\sum_{a^{'}\in\mtcalA}\mathbf{x}\left(s_{i}',a'\right)\pi_{\theta_{n}}\left(a'\mid s_{i}'\right)$,
and $\pi_{\theta_{n}}\left(a\mid s\right)$ is the policy for the
$n$-th user. Let $\mtbfR=\left[\mtbfr_{1},\cdots,\mtbfr_{N}\right]\in\mtbbR^{t\times N}$
store the reward of all $N$ users at all the $t$ time points. To
prevent the overfitting when $t$ is small at the beginning of online
RL learning, the $\ell_{2}$ norm based constraint is considered in
the objective as follows 
\begin{equation}
\mtbfw_{n}=\mhbfh_{n}=\arg\min_{\mtbfh_{n}}\left\Vert \mtbfX_{n}^{\intercal}\mtbfh_{n}-\left(\mtbfr_{n}+\gamma\mtbfY_{n}^{\intercal}\mtbfw_{n}\right)\right\Vert _{2}^{2}+\zeta_{c}\left\Vert \mtbfh_{n}\right\Vert _{2}^{2}
\end{equation}
$\mtfor\ n\in\left\{ 1,\cdots,N\right\} $. The LSTD$Q$ provides
a closed-form solution 
\begin{equation}
\mhbfw_{\theta_{n}}=\left[\mtbfX_{n}\left(\mtbfX_{n}-\gamma\mtbfY_{n}\right)^{\intercal}+\zeta\mtbfI\right]^{-1}\mtbfX_{n}\mtbfr_{n},\label{eq:LSTDQ_4_wn}
\end{equation}
for $\left\{ n\right\} _{n=1}^{N}$, where $\mhbfw_{\theta_{n}}$
is a function of the policy parameter $\theta_{n}$.

\subsection{The actor updating for policy improvement \label{sub:ActorUpdating_fmincon}}

In mHealth, the reference distribution of states $d_{\text{ref}}\left(s\right)$
is unknown and hard to estimate due to the lack of samples. We set
$d_{\text{ref}}\left(s\right)$ as the empirical distribution of states.
Accordingly, the observations in the trajectory, i.e. $\mtcalD_{n}$,
are used to form the objective for the actor updating $\widehat{\theta}_{n}={\displaystyle \arg\max_{\theta_{n}}\widehat{J}\left(\theta_{n}\right)}$,
where 
\begin{equation}
\widehat{J}\left(\theta_{n}\right)=\frac{1}{\left|\mtcalD_{n}\right|}\sum_{s_{i}\in\mtcalD_{n}}\sum_{a\in\mtcalA}Q\left(s_{i},a;\mhbfw_{\theta_{n}}\right)\pi_{\theta_{n}}\!\left(a|s_{i}\right)-\frac{\zeta_{a}}{2}\left\Vert \theta_{n}\right\Vert _{2}^{2}\label{eq:Actor_objective_trajectory}
\end{equation}
$\mtfor\ n\in\left\{ 1,\cdots,N\right\} $. Here $\left\Vert \theta_{n}\right\Vert _{2}^{2}$
is the constraint to make\ \eqref{eq:Actor_objective_trajectory}
a well-posed problem and $\zeta_{a}$ is the tuning parameter that
controls the strength of the smooth penalization\ \cite{huitian_2014_NIPS_ActCriticBandit4JITAI}.
We use $\widehat{J}\left(\theta_{n}\right)$ rather than $J\left(\theta_{n}\right)$
in\ \eqref{eq:Actor_objective_trajectory} to indicate that the objective
function for the actor updating is defined based on the $Q$-value
estimation.

Since the critic updating results in a closed-form solution\ \eqref{eq:LSTDQ_4_wn},
we could substitute the expression\ \eqref{eq:LSTDQ_4_wn} into the
objective for the actor updating\ \eqref{eq:Actor_objective_trajectory}.
Such case, however, leads to a very complex optimization problem.
In the case of large feature space, one can recursively update $\mhbfw_{\theta}$
and $\widehat{\theta}_{n}$ to reduce the computational cost.

\section{Network Cohesion based online Actor-Critic RL}

It is a famous phenomenon observed in lots of social behavior studies\ \cite{Haynie_2001_AJOS_NetworkStructure,Fujimoto_2012_SocialScience_NetworkInfluence}
that people are widely connected in a network and linked users tend
to have similar behaviors. Advances in social media help a lot to
record the relational information among users, which ensures the availability
of network information for health-related studies. Besides, individuals
are widely connected due to the similar features, such as age, gender,
race, religion, education level, work, income, other socioeconomic
status, medical records and genetics features etc\ \cite{Tianxi_2016_CORR_PredictModels4NetworkLinkedData}.
However, for simple study, current online methods for the mHealth
simply assume that users are completely different; they share no information
among users and learn a separate RL for each user by only using his
or her data. Such assumption works well in the ideal condition where
the sample drawn from each user is large in size to support the separate
online learning. However, though the data for all users is abundant,
the data for each user is limited in size. For example at the beginning
of online learning, there are $t=5$ tuples, which is hardly enough
to support a separate learning and likely to result in unstable policies.
From the perspective of optimization, the problem of lack of samples
badly affects the actor-critic updating not only at the beginning
of online learning but also along the whole learning process. This
is because the actor-critic objective functions are non-convex and
nonlinear; the bad solution at the beginning of online learning would
bias the optimization to sub-optimal directions. Besides, the policy
achieved at the early stage of online learning is of bad user experience,
which is likely for the users to be inactive with or even to abandon
the mHealth.

Different from current methods, we consider the phenomenon that a
user is similar to some (but not all) users, and similar users behave
similar but not completely identical to each other. To this end, we
propose a cohesion-based online RL method for the mHealth study. We
aim to understand how to share information across similar users in
order to improve the performance.

\subsection{Construct the network cohesion by using the warm start trajectory
(WST)}

\label{sub:ConstructCohesionNetwork} We assume there is an undirected
network cohesion connecting similar users, i.e. $\mtcalG=\left(\mtcalV,\mtcalE\right)$,
where $\mtcalV=\left\{ 1,2,\cdots,N\right\} $ is the set of nodes
(representing users) and $\mtcalE\subset\mtcalV\times\mtcalV$ is
the edge set. Altough advanced social medias, like Facebook, Twitter
and Linkedin, could provide us ith various network information, they
are not designed for the mHealth. There is noisy and misleading relational
information in the network for mHealth\ WS\cite{Bianchi_2013_NIPS_GangOfBandit,guangliangCheng_2016_JStars_robustHyperClassification,Alexandra_2016_AISTATS_GraphBandits,Claudio_2014_ICML_olineClusteringBandits,guangliangCheng_2016_TGRSL,fyzhu_2014_TIP_DgS_NMF,guangliangCheng_2014_ICIP,fyzhu_2014_AAAI_ARSS,guangliangCheng_2015_ICIP,yingWang_2015_TIP_RobustUnmixing,xiaoping_2017_ICASSP,guangliangCheng_2016_neurocomputing,fyzhu_2014_JSTSP_RRLbS}.
Thus, we want to learn the network cohesion intentionally for the
mHealth by measuring the similarities between the related behaviors
of users.

In RL, the MDP provides a mathematical tool to describe the property
of users in a specific mHealth study\footnote{The MDPs of one user on two diverse mHealth studies should be very
different; for example, the MDP in the HeartSteps study\ \cite{Walter_2015_Significance_RandomTrialForFitbitGeneration}
for one user should be different from that in the alcohol control\ \cite{Gustafson_2014_JAMA_drinking,Witkiewitz_2014_JAB_drinkingSmoking}
study.}. By measuring the similarities among the users' MDPs , we could learn
the network cohesion targeted to that mHealth study. However, the
MDP models are unknown to the RL problem. Instead, the warm start
trajectories (WSTs) of all the $N$ users are available, which provide
the observation of users. Thus, we use the WSTs for the graph learning,
i.e. $\mtcalD^{\left(0\right)}=\left\{ \mtcalD_{n}^{\left(0\right)}\mid n=1,\cdots,N\right\} $,
where $\mtcalD_{n}^{\left(0\right)}=\left\{ \left(s_{i,n},a_{i,n,}r_{i,n}\right)\right\} _{i=1}^{T_{0}}$
is the WST for the $n$-th user. Since an MDP consists of the state
transistion and immediate reward model, the feature for the cohesion
network learning is constructed by stacking the states and rewards
in the WST as follows 
\begin{equation}
\mtbfv_{n}=\left[s_{1,n}^{\intercal},r_{1,n},\cdots,s_{T_{0},n}^{\intercal},r_{T_{0},n}\right]^{\intercal}\in\mtbbR^{pT_{0}+T_{0}},\label{eq:feature_4_GraphLearning}
\end{equation}
for $n\in\left\{ 1,\cdots,N\right\} $. Note that the action or policy
is not part of an MDP. To reduce the influence of random actions in
the WST, we get rid of the temporal order by sorting all the elements
in $\mtbfv_{n}$\ \eqref{eq:feature_4_GraphLearning}. Then the benchmark
method, i.e. $K$-nearest neighbor ($K$NN), is used to learn the
neighboring information among users 
\begin{equation}
c_{ij}=\begin{cases}
1, & \mtif\ \mtbfv_{i}\in\mtcalN\left(\mtbfv_{j}\right)\ \text{or}\ \mtbfv_{j}\in\mtcalN\left(\mtbfv_{i}\right)\\
0, & \text{otherwise}
\end{cases},\label{eq:KNN_4_GraphLearning}
\end{equation}
where $\mtbfv_{i}\in\mtcalN\left(\mtbfv_{j}\right)$ indicates that
$i$-th node is the $K$NN of the $j$-th node\ \cite{Luxburg_2007_SC_SpectralClustering};\ \eqref{eq:KNN_4_GraphLearning}
is an undirected Graph. The value of $K$ controls how widely the
users are connected. A large $K$ indicates a wide connection among
users and vice versa.

\subsection{Model of cohesion based Actor-Critic RL \label{sub:Objective_4_GraphRL}}

The underlying assumption throughout this paper is that if two users
are connected, their values and policies are constrained to be similar,
e.g. $\left\Vert \mtbfw_{i}-\mtbfw_{j}\right\Vert $ and $\left\Vert \theta_{i}-\theta_{j}\right\Vert $
are small if $i\leftrightarrow j$\ \cite{fyzhu_2014_JSTSP_RRLbS,fyzhu_2015_PhDthesis}.
With the network cohesion $\mtbfC=\left(c_{ij}\right)_{N\times N}$,
the objective function for the critic updating is formed as follows
{\small{}
\begin{align}
\mtbfW=\mhbfH & =\arg\min_{\mtbfH}\sum_{n=1}^{N}\sum_{\mtcalU_{i}\in\mtcalD_{n}}\left\Vert \mtbfx{}_{i}^{\intercal}\mtbfh_{n}-\left(r_{i}+\gamma\mtbfy{}_{i}^{\intercal}\mtbfw_{n}\right)\right\Vert _{2}^{2}\label{eq:CriticUpdat_Graph_theory}\\
\mtsubTo & \sum_{i,j=1}^{N}c_{ij}d\left(\mtbfh_{i},\mtbfh_{j}\right)\leq\delta_{1}\ \text{and}\ \sum_{i,j=1}^{N}c_{ij}d\left(\mtbfw_{i},\mtbfw_{j}\right)\leq\delta_{2},\nonumber 
\end{align}
}where $\mtbfH\!=\left[\mtbfh_{1},\cdots,\mtbfh_{N}\right]\in\mtbbR^{u\times N}$
and $\mtbfW=\left[\mtbfw_{1},\cdots,\mtbfw_{N}\right]\in\mtbbR^{u\times N}$
are designed matrices that consist of all the $N$ users' variables
(each column summarizes the unknown varibile of one user); $d\left(\mtbfh_{i},\mtbfh_{j}\right)$
is a distance measure between two vectors; usually we set $d\left(\cdot,\cdot\right)$
as the Euler distance. With the matrix notations in Section\ \ref{sub:CriticUpdating_LSTDQ},
we turn\ \eqref{eq:CriticUpdat_Graph_theory} into the following
two-level nested optimization problems
\begin{align}
\mhbfH\!=\arg\min_{\mtbfH} & \Bigg(\sum_{n=1}^{N}\left\Vert \mtbfX_{n}^{\intercal}\mtbfh_{n}-\left(\mtbfr_{n}+\gamma\mtbfY_{n}^{\intercal}\mtbfw_{n}\right)\right\Vert _{2}^{2}+\label{eq:criticUpdating_projection}\\
 & \quad\mu_{1}\sum_{i,j=1}^{N}c_{ij}\left\Vert \mtbfh_{i}-\mtbfh_{j}\right\Vert _{2}^{2}+\zeta_{1}\sum_{n=1}^{N}\left\Vert \mtbfh_{n}\right\Vert _{2}^{2}\Bigg),\nonumber 
\end{align}
\begin{align}
\mhbfW\!=\arg\min_{\mtbfW} & \Bigg(\sum_{n=1}^{N}\left\Vert \Phi_{n}\mtbfw_{n}-\Phi_{n}\mhbfh_{n}\right\Vert _{2}^{2}+\label{eq:criticUpdating_fixedPoint}\\
 & \quad\mu_{2}\!\sum_{i,j=1}^{N}c_{ij}\left\Vert \mtbfw_{i}-\mtbfw_{j}\right\Vert _{2}^{2}+\zeta_{2}\sum_{n=1}^{N}\left\Vert \mtbfw_{n}\right\Vert _{2}^{2}\Bigg),\nonumber 
\end{align}
where $\Phi_{n}$ is a designed matrix to facilitate the optimization
of\ \eqref{eq:criticUpdating_fixedPoint}. The 1st level\ \eqref{eq:criticUpdating_projection}
projects the Bellman image onto a linear space (we refer\ \eqref{eq:criticUpdating_projection}
as the projection step); the 2nd level\ \eqref{eq:criticUpdating_fixedPoint}
deals with the fixed point problem (i.e. the fixed-point step)\ \cite{Mohammad_2011_RARL_RegularizedLSTD_L1L2}.

The objective for the actor updating is defined as follows 
\begin{equation}
\left\{ \widehat{\theta}_{1},\cdots,\widehat{\theta}_{n},\cdots,\widehat{\theta}_{N}\right\} =\arg\max_{\left\{ \theta_{n}\right\} _{n=1}^{N}}\widehat{J}\left(\theta_{1},\cdots,\theta_{N}\right),\label{eq:objective_actorUpdate}
\end{equation}
where $\Theta=\left[\theta_{1},\cdots,\theta_{N}\right]$, $Q\left(s_{i},a;\widehat{\mathbf{w}}_{\theta_{n}}\right)=\mathbf{x}\left(s_{i},a\right)^{T}\widehat{\mathbf{w}}_{\theta_{n}}$
is the estimated value for the $n$-th policy $\pi_{\theta_{n}}$
and 
\begin{align}
\widehat{J}\left(\Theta\right)\!= & \sum_{n=1}^{N}\left(\frac{1}{\left|\mtcalD_{n}\right|}\sum_{\mtcalU_{i}\in\mtcalD_{n}}\sum_{a\in\mtcalA}Q\left(s_{i},a;\mathbf{\mhbfw}_{\theta_{n}}\right)\pi_{\theta_{n}}\!\left(a|s_{i}\right)\right)\nonumber \\
 & \ -\frac{\mu_{3}}{2}\sum_{i,j=1}^{N}c_{ij}\left\Vert \theta_{i}-\theta_{j}\right\Vert _{2}^{2}-\frac{\zeta_{3}}{2}\sum_{n=1}^{N}\left\Vert \theta_{n}\right\Vert _{2}^{2}.\label{eq:actorUpdating_GraphRL}
\end{align}
Although we are able to obtain a closed-form solution for the critic
updating\ \eqref{eq:CriticUpdat_Graph_theory}, to reduce the computational
costs, we substitute the solution in value for $\left\{ \mhbfw_{n}\right\} _{n=1}^{N}$
rather than the closed-form expression of $\left\{ \mhbfw_{\theta_{n}}\right\} _{n=1}^{N}$
into the actor updating. The actor updating algorithm performs the
maximization of\ \eqref{eq:actorUpdating_GraphRL} over $\Theta$,
which is computed via the Sequential Quadratic Programming (SQP) algorithm.
We use the implementation of SQP with finite-difference approximation
to the gradient in the \textsc{fmincon} function of Matlab.

In the objectives\ \eqref{eq:criticUpdating_projection}, \eqref{eq:criticUpdating_fixedPoint}
and \eqref{eq:actorUpdating_GraphRL}, $\mu_{1}$, $\mu_{2}$ and
$\mu_{3}$ are the tuning parmaters to control the strength of the
network cohesion constraints. It is an advantage of our methods over
the network based bandit\ \cite{Bianchi_2013_NIPS_GangOfBandit}.
When $\mu_{1},\mu_{2},\mu_{3}\rightarrow\infty$, the connected users
are enforced to have identical values and policies. When $\mu_{1},\mu_{2},\mu_{3}=0$,
there is no network cohesion constraint. In such case, our method
is equivalent to the separate online RL method. Compared with the
Separate-RL, the model complexity of our methods is reduced since
their parameter domain is constrained via the network cohesion regularization.
Such case ensures our methods to work well when the sample size is
small. However, the optimization of our method is much more complex
than that of the separate-RL. The updating rules of all the users
are independent with each other in the separate-RL; while in our method,
the optimization of all the users is all coupled together. In the
following section, two actor-critic RL algorithms are proposed to
deal the objectives\ \eqref{eq:criticUpdating_projection} and \eqref{eq:criticUpdating_fixedPoint}.

\begin{algorithm}[t]
\caption{{\small{}Two online actor-critic algorithms for the Cohesion-RL\label{alg:2_actor_critic_algorithms_4_GraphRL}}}
\textbf{Input}: $T,T_{0},\mu_{\left\{ 1,2,3\right\} },\zeta_{\left\{ 1,2,3\right\} },nAlg$
(i.e. the algorithm index).

\begin{algorithmic}[1] 

\STATE Initialize the states $\left(s_{t,n}\right)_{p\times N}$,
where $t=0$, and the policy parameters $\Theta=\left[\theta_{1},\cdots,\theta_{N}\right]\in\mtbbR^{m\times N}$
for $N$ users. 

\FOR{ $n=1,\cdots,N$ }

\FOR{ $t=1,\cdots,T$ }

\STATE At time point $t,$ observe context $s_{t,n}$ for the $n$-th
user.

\STATE Draw an action $a_{t,n}$ according to the policy $\pi_{\widehat{\theta}_{n}}(a|s_{t,n}).$

\STATE Observe an immediate reward $r_{t,n}$.

\ENDFOR

\IF{ $t=T_{0}$}

\STATE Construct the network cohesion via\ \eqref{eq:feature_4_GraphLearning},$\ $\eqref{eq:KNN_4_GraphLearning}.

\ELSIF{ $t\geq T_{0}$ } 

\STATE Data preparation and feature construction\ \eqref{eq:featureConstruction_4_value}
for the critic update and the actor update.

\IF{ $nAlg=1$ } 

\STATE Critic update to learn $\mhbfW_{t}$ (value parameter) via\ \eqref{eq:CriticUpdatingRule_=00003D0000231_GraphRL}. 

\ELSIF{$nAlg=2$ }

\STATE Critic update to learn $\mhbfW_{t}$ (value parameter) via\ \eqref{eq:CriticUpdatingRule_=00003D0000232_GraphRL}.

\ENDIF

\STATE Actor update to learn $\widehat{\Theta}_{t}$ (policy paramter)
via\ \eqref{eq:objective_actorUpdate}.

\ENDIF

\ENDFOR

\end{algorithmic} 

\textbf{Output}: the policy for $N$ users, i.e. $\pi_{\widehat{\theta}_{n}}\left(a\mid s\right)$,
for $\left\{ n\right\} _{n=1}^{N}$. 
\end{algorithm}

\section{Algorithm\#1 for the Critic update}

\subsection{Updating Rules for the Projection Step\ \eqref{eq:criticUpdating_projection}
\label{sub:UpdatingRule4ProjectionStep}}

We first discuss how to minimize the objective for the projection
step. The objective is 
\begin{equation}
J\!=\!\sum_{n=1}^{N}\!\left\Vert \mtbfX_{n}^{\intercal}\mtbfh_{n}\!-\!\left(\mtbfr_{n}\!+\!\gamma\mtbfY_{n}^{\intercal}\mtbfw_{n}\right)\right\Vert _{2}^{2}+\mu_{1}\mttrace\left(\mtbfH\mtbfL\mtbfH^{\intercal}\right)\!+\!\zeta_{1}\left\Vert \mtbfH\right\Vert _{F}^{2},\label{eq:critic_obj_Graph}
\end{equation}
where $\left\Vert \cdot\right\Vert _{F}^{2}$ is the Frobenius norm
of a matrix, 
\begin{equation}
{\displaystyle \mttrace\left(\mtbfH\mtbfL\mtbfH^{\intercal}\right)=\sum_{i=1}^{N}\sum_{j=1}^{N}c_{ij}\left\Vert \mtbfh_{i}-\mtbfh_{j}\right\Vert _{2}^{2}}\label{eq:Deviation_Graph_Constraint}
\end{equation}
and $\mtbfL=\mtbfD-\mtbfC\in\mtbbR^{N\times N}$ is a graph laplacian;
$\mtbfD$ is a diagonal matrix whose elements are column (or row,
as $\mtbfC$ is a symmetric matrix) sums of $\mtbfC$, i.e. $d_{ii}=\sum_{i}c_{ij}$.
The partial derivative of $J_{1}$, i.e. the 1st term in\ \eqref{eq:critic_obj_Graph},
with respect to $\mtbfh_{n}$ is 
\begin{equation}
\frac{\partial J_{1}}{\partial\mtbfh_{n}}=2\mtbfX_{n}\mtbfX_{n}^{\intercal}\mtbfh_{n}-2\mtbfX_{n}\mtbfr_{n}-2\gamma\mtbfX_{n}\mtbfY_{n}^{\intercal}\mtbfw_{n}.
\end{equation}
Summarizing the partial derivatives with respect to all the variables
in $\mtbfH=\left(\mtbfh_{1},\cdots,\mtbfh_{N}\right)\in\mtbbR^{K\times N}$,
we have 
\begin{align}
\frac{\partial J_{1}}{\partial\mtvec\left(\mtbfH\right)}=\  & 2\left(\sum_{n=1}^{N}\mtbfE_{n}\otimes\left(\mtbfX_{n}\mtbfX_{n}^{\intercal}\right)\right)\mtvec\left(\mtbfH\right)\nonumber \\
 & -2\left(\sum_{n=1}^{N}\mtbfE_{n}\otimes\mtbfX_{n}\right)\mtvec\left(\mtbfR\right)\nonumber \\
 & -2\gamma\left(\sum_{n=1}^{N}\mtbfE_{n}\otimes\left(\mtbfX_{n}\mtbfY_{n}^{\intercal}\right)\right)\mtvec\left(\mtbfW\right)
\end{align}
where $\mtvec\left(\mtbfH\right)\!=\!\left[\mtbfh_{1}^{\intercal},\cdots,\mtbfh_{N}^{\intercal}\right]^{\intercal}\!\in\!\mtbbR^{uN}$
is the vectorization process for a matrix; $\mtbfE_{n}=\mtdiag\left(0,\cdots,1,\cdots,0\right)\in\mtbbR^{N\times N}$
is a diagonal matrix with the $n$-th diagonal element equal to 1,
all the other equal to zero; $\otimes$ indicates the Kronecker product
between two matrices resulting in a block matrix. Let $\mtbfF_{1}=\sum_{n}\mtbfE_{n}\otimes\left(\mtbfX_{n}\mtbfX_{n}^{\intercal}\right)$,
$\mtbfF_{2}=\sum_{n}\mtbfE_{n}\otimes\mtbfX_{n}$ and $\mtbfF_{3}=\sum_{n}\mtbfE_{n}\otimes\left(\mtbfX_{n}\mtbfY_{n}^{\intercal}\right)$.
We have a simpler formulation for the $\partial J_{1}/\partial\mtvec\left(\mtbfH\right)$
as follows 
\begin{equation}
\frac{\partial J_{1}}{\partial\mtvec\left(\mtbfH\right)}=2\mtbfF_{1}\mtvec\left(\mtbfH\right)-2\left[\mtbfF_{2}\mtvec\left(\mtbfR\right)+\gamma\mtbfF_{3}\mtvec\left(\mtbfW\right)\right].
\end{equation}
The partial derivatives of the 2nd term in\ \eqref{eq:critic_obj_Graph},
i.e. $J_{2}=\mu_{1}\mttrace\left(\mtbfH\mtbfL\mtbfH^{\intercal}\right)+\zeta_{1}\left\Vert \mtbfH\right\Vert _{F}^{2}$,
with respect to $\mtbfH$ is 
\[
\frac{\partial J_{2}}{\partial\mtbfH}=2\mu_{1}\mtbfH\mtbfL+2\zeta_{1}\mtbfH.
\]
According to the Encapsulating Sum\ \cite{Petersen_2012_MatrixCookbook},
we have 
\begin{equation}
\frac{\partial J_{2}}{\partial\mtvec\left(\mtbfH\right)}=2\left[\left(\mu_{1}\mtbfL^{\intercal}+\zeta_{1}\mtbfI_{N}\right)\otimes\mtbfI_{u}\right]\mtvec\left(\mtbfH\right)\label{eq:gradient_with_Graph_constraint}
\end{equation}
where $\mtbfI_{N}\in\mtbbR^{N\times N}$ and $\mtbfI_{u}\in\mtbbR^{u\times u}$
are identical matrices. Setting the gradient of $J$ in\ \eqref{eq:critic_obj_Graph}
with respect to $\mtvec\left(\mtbfH\right)$ to zero gives the closed-form
solution for the projection step as follows 
\begin{equation}
\mtvec\left(\mhbfH\right)=\left[\mtbfF_{1}+\mtbfL_{\otimes}\left(\mu_{1},\zeta_{1}\right)\right]^{-1}\left[\mtbfF_{2}\mtvec\left(\mtbfR\right)+\gamma\mtbfF_{3}\mtvec\left(\mtbfW\right)\right],\label{eq:closedFormSolution_4_projectionStep}
\end{equation}
where $\mtbfL_{\otimes}\left(\mu_{1},\zeta_{1}\right)=\left(\mu_{1}\mtbfL^{\intercal}+\zeta_{1}\mtbfI_{N}\right)\otimes\mtbfI_{u}\in\mtbbR^{uN\times uN}$.

\subsection{Updating Rules for the Fixed Point step\ \eqref{eq:criticUpdating_fixedPoint}}

Considering the 1st term in the fixed point step\ \eqref{eq:criticUpdating_fixedPoint}
gives 

\begin{align*}
O_{1} & =\sum_{n}\left\Vert \Phi_{n}\mtbfw_{n}-\Phi_{n}\mhbfh_{n}\right\Vert _{2}^{2}=\left\Vert \Phi_{\otimes}\mtvec\left(\mtbfW\right)-\Phi_{\otimes}\mtvec\left(\mhbfH\right)\right\Vert _{2}^{2}
\end{align*}
where $\Phi_{\otimes}=\left(\sum_{n=1}^{N}\mtbfE_{n}\otimes\Phi_{n}\right)$.
To facilitate the optimization, we design $\left\{ \Phi_{n}\right\} _{n=1}^{N}$
to let $\Phi_{\otimes}=\mtbfF_{1}+\mtbfL_{\otimes}\left(\mu_{1},\zeta_{1}\right)$,
which leads to 
\begin{align*}
O_{1} & =\left\Vert \Phi_{\otimes}\mtvec\left(\mtbfW\right)-\Phi_{\otimes}\mtvec\left(\mhbfH\right)\right\Vert _{2}^{2}\\
 & =\left\Vert \Phi_{\otimes}\mtvec\left(\mtbfW\right)-\left[\mtbfF_{2}\mtvec\left(\mtbfR\right)+\gamma\mtbfF_{3}\mtvec\left(\mtbfW\right)\right]\right\Vert \\
 & =\left\Vert \left(\Phi_{\otimes}-\gamma\mtbfF_{3}\right)\mtvec\left(\mtbfW\right)-\mtbfF_{2}\mtvec\left(\mtbfR\right)\right\Vert _{2}^{2},
\end{align*}
and finally results in an easy solution for the critic updating\ \eqref{eq:CriticUpdatingRule_=00003D0000231_GraphRL}
(cf. Theorem\ \ref{thm:B_sysmmetirc_positive_definite}). Letting
$\mtbfP=\Phi_{\otimes}-\gamma\mtbfF_{3}$, we have 
\[
O_{1}=\left\Vert \mtbfP\mtvec\left(\mtbfW\right)-\mtbfF_{2}\mtvec\left(\mtbfR\right)\right\Vert _{2}^{2}.
\]
The partial derivative of $O_{1}$ with respect to $\mtvec\left(\mtbfW\right)$
is 
\[
\frac{\partial O_{1}}{\partial\mtvec\left(\mtbfW\right)}=2\mtbfP^{\intercal}\mtbfP\mtvec\left(\mtbfW\right)-2\mtbfP^{\intercal}\mtbfF_{2}\mtvec\left(\mtbfR\right).
\]
Considering the partial derivative of the cohesion constraint and
the Frobenius norm based smooth constraint with respect to $\mtvec\left(\mtbfH\right)$,
and setting the overll partial derivative to zero, i.e. $\partial O/\partial\mtvec\left(\mtbfW\right)=\mtbfzero$,
we can obtain the following closed-form solution 
\begin{equation}
\mtvec\left(\mtbfW^{*}\right)=\left[\mtbfP^{\intercal}\mtbfP+\mtbfL_{\otimes}\left(\mu_{2},\zeta_{2}\right)\right]^{-1}\mtbfP^{\intercal}\mtbfF_{2}\mtvec\left(\mtbfR\right),\label{eq:CriticUpdatingRule_=00003D0000231_GraphRL}
\end{equation}
where $\mtbfL_{\otimes}\left(\mu_{2},\zeta_{2}\right)=\left(\mu_{2}\mtbfL^{\intercal}+\zeta_{2}\mtbfI_{N}\right)\otimes\mtbfI_{u}$.
\begin{thm}
\label{thm:B_sysmmetirc_positive_definite} $\mtbfB=\mtbfP^{\intercal}\mtbfP+\mtbfL_{\otimes}\left(\mu_{2},\zeta_{2}\right)$
is a symmetric and positive definite matrix, which leads to an easy
critic updating rule in\ \eqref{eq:CriticUpdatingRule_=00003D0000231_GraphRL}. \end{thm}
\begin{lem}
\label{lem:KroneckerProduct_eigenvalues} Suppose that $\mtbfA\in\mtbbR^{n\times n}$
and $\mtbfB\in\mtbbR^{m\times m}$ are square matrices. Let $\lambda_{1},\cdots,\lambda_{n}$
be the eigenvalues of $\mtbfA$ and $\nu_{1},\cdots,\nu_{m}$ be those
of $\mtbfB$. Then the eigenvalues of $\mtbfA\otimes\mtbfB$\ \cite{Langville_2004_JCAM_KroneckerProduct},
where $\otimes$ is the Kronecker Product, are 
\[
\lambda_{i}\nu_{j},\qquad\mtfor\ i\in\left\{ 1,\cdots,n\right\} ,\ j\in\left\{ 1,\cdots,m\right\} .
\]

\end{lem}
\begin{table*}[tbh]
\begin{centering}
\caption{The performance of 3 online RLs vs. the rising trajectory length $T\in\{50,150\}$,
for the experiment setting (\textbf{S1}). \label{tab:_ExpSetting_S1}}
\begin{tabular}{|c||c|c|c||c|c|c|}
\hline 
\multirow{2}{*}{$\gamma$ } &
\multicolumn{3}{c|||}{Average reward when $T=50$ } &
\multicolumn{3}{c|}{Average reward when $T=150$}\tabularnewline
\cline{2-7} 
 & Separate-RL  &
Cohesion-RL\#1  &
Cohesion-RL\#2  &
Separate-RL  &
Cohesion-RL\#1  &
Cohesion-RL\#2\tabularnewline
\hline 
$0$  &
1238.4$\pm$81.6  &
\textcolor{blue}{\emph{1332.3$\pm$57.0}}  &
\textbf{1394.4$\pm$68.5}  &
1239.5$\pm$83.9  &
\textcolor{blue}{\emph{1342.1$\pm$59.7}}  &
\textbf{1397.4$\pm$68.6}\tabularnewline
$0.2$  &
1272.9$\pm$83.7  &
\textcolor{blue}{\emph{1376.1$\pm$55.8}}  &
\textbf{1428.1$\pm$64.1}  &
1279.5$\pm$83.4  &
\textcolor{blue}{\emph{1386.4$\pm$56.0}}  &
\textbf{1429.2$\pm$64.3}\tabularnewline
$0.4$  &
1286.7$\pm$85.3  &
\textcolor{blue}{\emph{1429.3$\pm$53.0}}  &
\textbf{1472.4$\pm$58.7}  &
1316.4$\pm$77.5  &
\textcolor{blue}{\emph{1436.2$\pm$55.6}}  &
\textbf{1472.4$\pm$59.0}\tabularnewline
$0.6$  &
1346.3$\pm$75.9  &
\textcolor{blue}{\emph{1488.4$\pm$55.5}}  &
\textbf{1505.1$\pm$55.4}  &
1388.5$\pm$70.8  &
\textcolor{blue}{\emph{1502.6$\pm$52.4}}  &
\textbf{1515.3$\pm$55.9}\tabularnewline
$0.8$  &
1373.9$\pm$66.2  &
\textcolor{blue}{\emph{1550.8$\pm$52.6}}  &
\textbf{1556.7$\pm$54.0}  &
1440.7$\pm$63.5  &
\textcolor{blue}{\emph{1560.4$\pm$53.1}}  &
\textbf{1570.0$\pm$54.0}\tabularnewline
$0.95$  &
\multicolumn{1}{c|}{1265.2$\pm$78.3} &
\textcolor{blue}{\emph{1542.7$\pm$50.7}}  &
\textbf{1556.5$\pm$49.6}  &
1315.2$\pm$68.6  &
\textcolor{blue}{\emph{1570.0$\pm$50.3}}  &
\textbf{1577.8$\pm$51.3}\tabularnewline
\hline 
Avg.  &
1297.2  &
\textcolor{blue}{\emph{1453.3}}  &
\textbf{1485.5}  &
1330.0  &
\textcolor{blue}{\emph{1466.3}}  &
\textbf{1493.7 }\tabularnewline
\hline 
\end{tabular}
\par\end{centering}

\begin{centering}
\caption{The performance of 3 online RLs vs. the warm start trajectory length
$T_{0}\in\{5,20\}$, for the experiment setting (\textbf{S2}). \label{tab:_ExpSetting_S2}}
\begin{tabular}{|c||c|c|c||c|c|c|}
\hline 
\multirow{2}{*}{$\gamma$ } &
\multicolumn{3}{c||}{Average reward when $T_{0}=5$ } &
\multicolumn{3}{c|}{Average reward when $T_{0}=20$ }\tabularnewline
\cline{2-7} 
 & Separate-RL  &
Cohesion-RL\#1  &
Cohesion-RL\#2  &
Separate-RL  &
Cohesion-RL\#1  &
Cohesion-RL\#2\tabularnewline
\hline 
$0$  &
1194.0$\pm$86.8  &
\textcolor{blue}{\emph{1324.7$\pm$60.0}}  &
\textbf{1396.7$\pm$68.5}  &
\textcolor{blue}{\emph{1380.8$\pm$68.7}}  &
1358.9$\pm$63.6  &
\textbf{1399.1$\pm$68.0}\tabularnewline
$0.2$  &
1183.8$\pm$87.8  &
\textcolor{blue}{\emph{1380.5$\pm$52.0}}  &
\textbf{1427.4$\pm$64.5}  &
\textcolor{blue}{\emph{1410.1$\pm$66.8}}  &
1408.0$\pm$57.8  &
\textbf{1430.2$\pm$63.6}\tabularnewline
$0.4$  &
1200.4$\pm$81.1  &
\textcolor{blue}{\emph{1433.2$\pm$54.2}}  &
\textbf{1469.2$\pm$58.2}  &
1423.7$\pm$63.9  &
\textcolor{blue}{\emph{1453.9$\pm$55.4}}  &
\textbf{1471.3$\pm$59.3}\tabularnewline
$0.6$  &
1254.1$\pm$75.7  &
\textcolor{blue}{\emph{1487.9$\pm$51.2}}  &
\textbf{1515.2$\pm$55.0}  &
1463.6$\pm$58.1  &
\textcolor{blue}{\emph{1505.7$\pm$53.8}}  &
\textbf{1516.8$\pm$56.1}\tabularnewline
$0.8$  &
1291.8$\pm$76.8  &
\textcolor{blue}{\emph{1532.7$\pm$48.6}}  &
\textbf{1562.3$\pm$53.8}  &
1519.8$\pm$53.9  &
\textcolor{blue}{\emph{1541.3$\pm$52.4}}  &
\textbf{1552.9$\pm$54.0}\tabularnewline
$0.95$  &
\multicolumn{1}{c|}{1245.1$\pm$81.5} &
\textcolor{blue}{\emph{1554.7$\pm$49.4}}  &
\textbf{1568.5$\pm$51.0}  &
1434.4$\pm$53.3  &
\textcolor{blue}{\emph{1564.1$\pm$49.2}}  &
\textbf{1574.6$\pm$50.0}\tabularnewline
\hline 
Avg.  &
1228.2  &
\textcolor{blue}{\emph{1452.3}}  &
\textbf{1489.9}  &
1438.7  &
\textcolor{blue}{\emph{1472.0}}  &
\textbf{1490.8 }\tabularnewline
\hline 
\end{tabular}
\par\end{centering}

\emph{The value of $\gamma$ specifies different RL methods: (a) $\gamma=0$
means the contextual bandit\ \cite{huitian_2014_NIPS_ActCriticBandit4JITAI};
(b) $0<\gamma<1$ is the discounted reward RL, which is first compared
in the online actor-critic setting for mHealth. In each comparision,
the }\textbf{\emph{bold value}}\emph{ is the best, and the }\textit{\textcolor{blue}{\emph{blue
itatlic value}}}\emph{ is the 2nd best.} 
\end{table*}

\section{Algorithm\#2 for the Critic update}

In this section, we provide another updating rule for the critic update
(i.e. policy improvement). Note that to prevent the overfitting when
the sample size is very small, the conventional LSTD$Q$ usually employs
the $\ell_{2}$ constraint on the variable $\mtbfH$ in the projection
step. They do not put the $\ell_{2}$ constraint on the fixed-point
variable $\mtbfW$\ \cite{Mohammad_2011_RARL_RegularizedLSTD_L1L2,AndrewNg_2009_ICML_RLsparity}.
Following this idea, we have a simpler objective function for the
critic update as 
\begin{align}
\mtbfw_{n} & =\mhbfh_{n}=\arg\min_{\mtbfh_{n}}\sum_{\mtcalU_{i}\in\mtcalD_{n}}\left\Vert \mtbfx{}_{i}^{\intercal}\mtbfh_{n}-\left(r_{i}+\gamma\mtbfy{}_{i}^{\intercal}\mtbfw_{n}\right)\right\Vert _{2}^{2},\\
 & \mtfor\ n\in\left\{ 1,\cdots,N\right\} \ \text{and}\ \mtsubTo\ \sum_{i,j=1}^{N}c_{ij}d\left(\mtbfh_{i},\mtbfh_{j}\right)\leq\delta_{1}.\nonumber 
\end{align}
According to the derivation in Section\ \ref{sub:UpdatingRule4ProjectionStep}
that considers the Frobenius norm based smooth constraint, the updating
rule for the projection step is\ \eqref{eq:closedFormSolution_4_projectionStep}.
In the fixed-point step, the objective is simply $\mtbfw_{n}=\mhbfh_{n}$
(i.e. a fixed-point problem), which leads to 
\[
\mtvec\left(\mtbfW\right)=\mtvec\left(\mhbfH\right).
\]
Thus, we have the closed-form solution for $\mtvec\left(\mtbfW\right)$
as follows 
\begin{equation}
\mtvec\left(\mhbfW\right)=\left[\mtbfF_{1}-\gamma\mtbfF_{3}+\mtbfL_{\otimes}\left(\mu_{1},\zeta_{1}\right)\right]^{-1}\mtbfF_{2}\mtvec\left(\mtbfR\right).\label{eq:CriticUpdatingRule_=00003D0000232_GraphRL}
\end{equation}
It is simpler than the 1st updating rule for the critic update\ \eqref{eq:CriticUpdatingRule_=00003D0000231_GraphRL}.

\section{Experiment Results}

\label{sec:Evaluation} We verify the proposed methods on the HeartSteps
dataset. It has two choices for an action, i.e. $\left\{ 0,1\right\} ,$
where $a=1$ means sending the positive intervention, while $a=0$
indicates no intervention\ \cite{SusanMurphy_2016_CORR_BatchOffPolicyAvgRwd}.
Specifically, the stochastic policy is assumed to be in the form $\pi_{\theta}\left(a\mid s\right)\!=\!\frac{\exp\left[-\theta^{\intercal}\phi\left(s,a\right)\right]}{\sum_{a'}\exp\left[-\theta^{\intercal}\phi\left(s,a'\right)\right]}$,
where $\theta\in\mtbbR^{m}$ is the unknown parameter and $\phi\left(\cdot,\cdot\right)$
is a feature process that combines the information in actions and
states, i.e. $\phi\left(s,a\right)=\left[as^{\intercal},a\right]^{\intercal}\in\mtbbR^{m}$.

\subsection{The HeartSteps Dataset}

To verify the performance of our method, we use a dataset from a mobile
health study, called HeartSteps\ \cite{Walter_2015_Significance_RandomTrialForFitbitGeneration},
to approximate the generative model. This is a 42-day mHealth intervention
that aims to increase the users' steps they take each day by providing
positive treatments (i.e. interventions), which are adapted to users'
ongoing status, such as suggesting users to take a walk after long
sitting\ \cite{Walter_2015_Significance_RandomTrialForFitbitGeneration},
or to do some exercises after work.

A trajectory of $T$ tuples $\mtcalD=\left\{ \left(s_{i},a_{i,}r_{i}\right)\mid i=1,\cdots,T\right\} $
are generated for each user\ \cite{SusanMurphy_2016_CORR_BatchOffPolicyAvgRwd,huitian_2014_NIPS_ActCriticBandit4JITAI}.
The initial state is drawn from the Gaussian distribution $S{}_{0}\sim\mtcalN_{p}\left\lbrace 0,\Sigma\right\rbrace $,
where $\Sigma$ is a $p\times p$ covariance matrix with pre-defined
elements. The action $a_{t}$ for $0\leq t\leq T_{0}$ is drawn from
the random policy, with a probability of $0.5$ to provide interventions,
i.e. $\mu\left(1\mid s\right)=0.5$ for all states $s$. Such process
is called drawing warm start trajectory (WST) via the micro-randomized
trials\ \cite{PengLiao_2015_Proposal_offPolicyRL,Walter_2015_Significance_RandomTrialForFitbitGeneration},
and $T_{0}$ is the length of the WST. When $t\geq T_{0}$, we start
the actor-critic updating, and the action is drawn from the learned
policy, i.e. $a_{t}\sim\pi_{\widehat{\theta}_{t}}\left(\cdot\mid s_{t}\right)$.
When $t\geq1$, the state and immediate reward are generated as follows
\begin{align}
S_{t,1}=\  & \beta_{1}S_{t-1,1}+\xi_{t,1},\nonumber \\
S_{t,2}=\  & \beta_{2}S_{t-1,2}+\beta_{3}A_{t-1}+\xi_{t,2},\label{eq:Dat=00003D0000231_stateTrans_cmp3}\\
S_{t,3}=\  & \beta_{4}S_{t-1,3}+\beta_{5}S_{t-1,3}A_{t-1}+\beta_{6}A_{t-1}+\xi_{t,3},\nonumber \\
S_{t,j}=\  & \beta_{7}S_{t-1,j}+\xi_{t,j},\qquad\mtfor\ j=4,\ldots,p\nonumber \\
R_{t}=\  & \beta_{14}\times[\beta_{8}+A_{t}\times(\beta_{9}+\beta_{10}S_{t,1}+\beta_{11}S_{t,2})\label{eq:Dat=00003D0000231_ImmediateRwd_cmp3}\\
 & +\beta_{12}S_{t,1}-\beta_{13}S_{t,3}+\varrho_{t}],\nonumber 
\end{align}
where $\bm{\beta}\!=\!\left\{ \beta_{i}\right\} _{i=1}^{14}$ is the
main parameter for the MDP and $-\beta_{13}S_{t,3}$ is the treatment
fatigue; $\left\{ \xi_{t,i}\right\} _{i=1}^{p}\sim\mtcalN\left(0,\sigma_{s}^{2}\right)$
is the noise in the state\ \eqref{eq:Dat=00003D0000231_stateTrans_cmp3}
and $\varrho_{t}\sim\mtcalN\left(0,\sigma_{r}^{2}\right)$ is the
noise in the reward model\ \eqref{eq:Dat=00003D0000231_ImmediateRwd_cmp3}.

As it is known to us, the individuals are generally more or less different
from each other, and each individual is similar to a part, but not
all, of the individuals. In the mHealth and RL study, an individual
is abstracted as an MDP, which is determined by the value of $\bm{\beta}$,
cf.\ \eqref{eq:Dat=00003D0000231_stateTrans_cmp3} and\ \eqref{eq:Dat=00003D0000231_ImmediateRwd_cmp3}.
To achieve a more practical dataset compared with\ \cite{SusanMurphy_2016_CORR_BatchOffPolicyAvgRwd,PengLiao_2015_Proposal_offPolicyRL,huitian_2014_NIPS_ActCriticBandit4JITAI},
we come up with a method to generate $N$ users (i.e. $\bm{\beta}$s)
that satisfy the above requirements in two steps: (a) manually design
$V$ basic $\bm{\beta}$s, i.e. $\left\{ \bm{\beta}_{v}^{\text{basic}}\mid v=1,\cdots,V\right\} $,
that are very different from each other; (b) a set of $N_{v}$ different
individuals (i.e. $\bm{\beta}$s) are generated for each $\bm{\beta}_{v}^{\text{basic}}$
via the following process $\bm{\beta}_{i}=\bm{\beta}_{v}^{\text{basic}}+\bm{\delta}_{i},\ \text{for}\ i\in\left\{ 1,2,\cdots,N_{v}\right\} $,
where $\bm{\delta}_{i}\sim\mtcalN\left(0,\sigma_{b}\mtbfI_{14}\right)$
is the noise in the MDPs and $\mtbfI_{14}\in\mtbbR^{14\times14}$
is an identity matrix. After such processing, the individuals are
all different from the others. The value of $\sigma_{b}$ specifies
how different the individuals are. In the experiments, the number
of groups is set as $V=3$ (each group has $N_{v}=15$ people, leading
to $N=45$ users involved in the experiment). The $\bm{\beta}^{\text{basic}}$'s
for the $V$ groups are set as follows 
\begin{align*}
\bm{\beta}_{1}^{\text{basic}}= & [0.40,0.25,0.35,0.65,0.10,0.50,0.22,\\
 & 2.00,0.15,0.20,0.32,0.10,0.45,800]\\
\bm{\beta}_{2}^{\text{basic}}= & [0.35,0.30,0.30,0.60,0.05,0.65,0.28,\\
 & 2.60,0.35,0.45,0.45,0.15,0.50,650]\\
\bm{\beta}_{3}^{\text{basic}}= & [0.20,0.50,0.20,0.62,0.06,0.52,0.27,\\
 & 3.00,0.15,0.15,0.50,0.16,0.70,450].
\end{align*}

\begin{figure*}[tbh]
\begin{centering}
\includegraphics[width=0.94\linewidth]{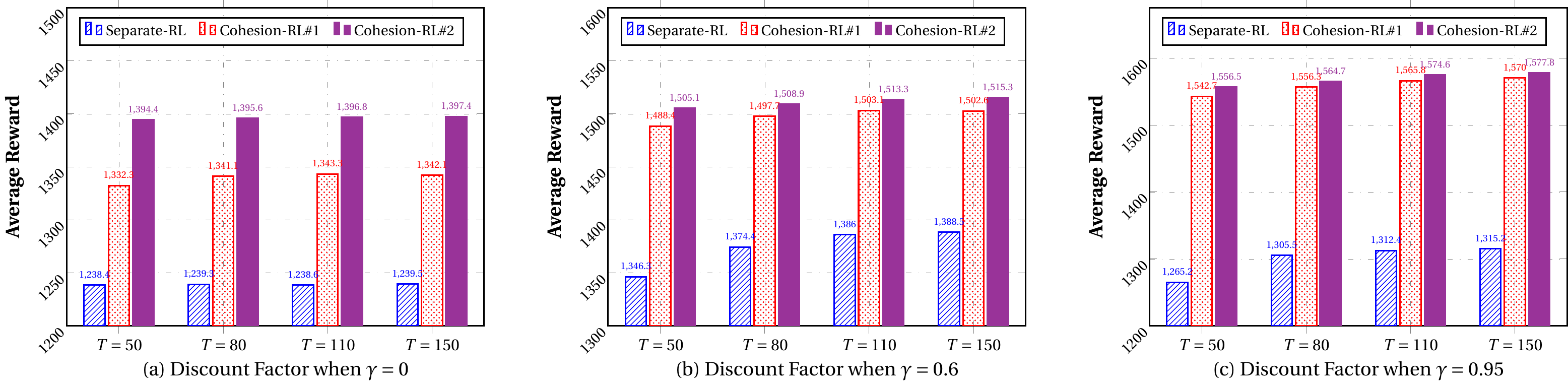}
\caption{Performance of three RL methods for experiment setting (\textbf{S1}).
Each sub-figure shows results under one $\gamma$ setting. \label{fig:_ExpSetting_S1}}

\par\end{centering}

\centering{}\includegraphics[width=0.94\linewidth]{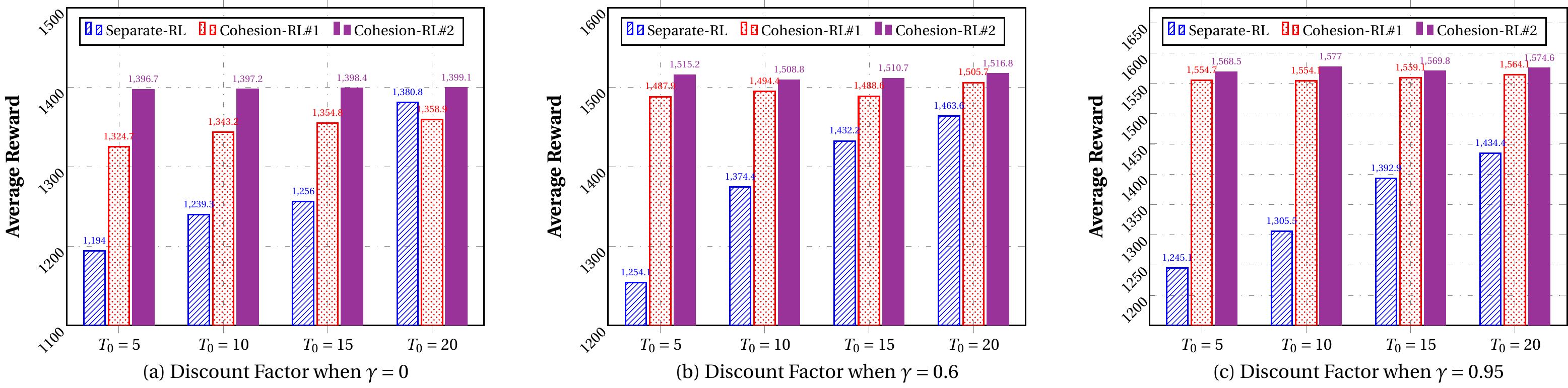}
\caption{Performance of three RL methods for experiment setting (\textbf{S2}).
Each sub-figure shows results under one $\gamma$ setting. \label{fig:_ExpSetting_S2}}
 
\end{figure*}

\subsection{\label{sub:ComparedMethods_ParameterSetting}Compared Methods and
Parameter Settings}

There are three online actor-critic RL methods for the comparison:
(a) Separate-RL, which is an extension of the online actor-critic
contextual bandit in\ \cite{huitian_2014_NIPS_ActCriticBandit4JITAI}
to the online actor-critic reinforcement learning. It learns a separate
RL policy for each user by only using his or her data. (b) Cohesion-RL\#1
is the first version of our method. (c) Cohesion-RL\#2 is the second
version of our method (cf. Algorithm\ \ref{alg:2_actor_critic_algorithms_4_GraphRL}
for detail). Specially, Cohesion-RL\#1 and Cohesion-RL\#2 share the
same actor updating. The difference between them is the different
critic updating rules that they employ.

The noises in the MDP are set as $\sigma_{s}\!=\!0.5$, $\sigma_{r}\!=\!1$
and $\sigma_{\beta}\!=\!0.05$. The state has dimension $p=3$ and
the policy feature has $m\!=\!4$ elements. We set the $\ell_{2}$
constraint in the Separate-RL as $\zeta_{a}=\zeta_{c}=0.1$. When
the cohesion constraint in our methods are too small ($10^{-4}$,
say), we need the $\ell_{2}$ constraint for the actor-critic updating
to avoid the overfitting, with the parameters as $\zeta_{1}=\zeta_{2}=\zeta_{3}=0.1$.
Otherwise, we set $\zeta_{1}=\zeta_{2}=\zeta_{3}\rightarrow0$. The
feature processing for the value estimation is $\mtbfx\left(s,a\right)=\left[1,s^{\intercal},a,s^{\intercal}a\right]^{\intercal}\in\mtbbR^{u},$
where $u=2p+2$, for all the compared methods. The feature for the
policy is processed as $\phi\left(s,a\right)=\left[as^{\intercal},a\right]^{\intercal}\in\mtbbR^{m}$
where $m=p+1$. We set $K=8$ for the $K$-NN based network cohesion
learning. If there is no special setting, the following three paremeters
are set as: (a) the trajectory length in mHealth is $T=80$, which
indicates that the online RL learning ends at $t=80$; (b) the length
of warm start trajectory is set as $T_{0}=10$; (c) to reduce the
number of parameters in the algorithm, the parameters for the cohesion
constraint in our methods are set as $\mu_{1}=0.1$, $\mu_{3}=\mu_{1}$
and $\mu_{2}=0.01\mu_{1}$.

\subsection{Evaluation Metrics}

We use the expectation of long run average reward (ElrAR) $\mathbb{E}\left[\eta^{\pi_{\widehat{\Theta}}}\right]$
to quantify the quality of the estimated policy $\pi_{\widehat{\Theta}}$
on a set of $N$=45 individuals. Here $\pi_{\widehat{\Theta}}$ summarizes
the policies for all the $45$ users, in which $\pi_{\widehat{\theta}_{n}}$
is the $n$-th user's policy. Intuitively, ElrAR measures how much
average reward in the long run we could totally get by using the learned
policy $\pi_{\widehat{\Theta}}$ on the testing users (i.e. MDPs),
for example measuring how much alcohol users have in a fixed time
period in the alcohol use study\ \cite{Gustafson_2014_JAMA_drinking,Witkiewitz_2014_JAB_drinkingSmoking}.
Specifically in the HeartSteps, ElrAR measures the average steps that
users take per day over a long time; a larger ElrAR corresponds to
a better performance. The average reward for the $n$-th user, i.e.
$\eta^{\pi_{\widehat{\theta}_{n}}}$, is calculated by averaging the
rewards over the last $4,000$ elements in a trajectory of $5,000$
tuples under the policy $\pi_{\widehat{\theta}_{n}}$, i.e. $\eta^{\pi_{\widehat{\theta}_{n}}}=\frac{1}{T-i}\sum_{j=i}^{T}\mtcalR\left(s_{j,n},a_{j,n}\sim\pi_{\widehat{\theta}_{n}}\right)$,
where $T=5000$ and $i=1000$. Then ElrAR $\mathbb{E}\left[\eta^{\pi_{\widehat{\Theta}}}\right]$
is approximated by averaging over the $45$ $\eta^{\pi_{\widehat{\theta}_{n}}}$'s,
i.e. $\mathbb{E}\left[\eta^{\pi_{\widehat{\Theta}}}\right]\approx\frac{1}{N}\sum_{n=1}^{N}\eta^{\pi_{\widehat{\theta}_{n}}}$.

\subsection{Comparisons in three experiment settings }

The following experiments are carried out to verify different aspects
of the three online actor-critic RL algorithms:

(\textbf{S1}) In this part, the trajectory length of all users ranges
as $T\!\in\left\{ 50,80,110,150\right\} $. The experiment results
are showed in Table\ \ref{tab:_ExpSetting_S1} and Fig.\ \ref{fig:_ExpSetting_S2}.
There are two sub-tables in Table\ \ref{tab:_ExpSetting_S1}; each
sub-table displays the ElrAR of three RL methods (i.e. Separate-RL,
Cohesion-RL\#1 and Cohesion-RL\#2 respectively) under six $\gamma$
settings; the last row shows the average ElrAR over the results of
all the six $\gamma$s. In Fig.\ \ref{fig:_ExpSetting_S2}, there
are three sub-figures; each sub-figure illustrates the results of
three methods under one $\gamma$ setting. As we shall see that the
performance of three methods generally increases as $T$ rises. The
performance of our RL methods, i.e. Cohesion-RL\#1 and Cohesion-RL\#2,
have an obvious advantage over the Separate-RL under all the parameters
settings in (\textbf{S1}). Besides, the advantage of our methods over
Separate-RL slowly decreases as $T$ rises. Compared with Separate-RL,
our methods averagely improve $156.0$ steps and $188.3$ steps when
$T=50$, and averagely improve $136.3$ steps and $163.7$ steps when
$T=150$.

(\textbf{S2}) In this part, the length of warm start trajectory ranges
as $T_{0}=\left\{ 5,10,15,20\right\} $, which indicates that the
RL methods wait longer and longer before starting the online learning.
The experiment results are summarized in Table\ \ref{tab:_ExpSetting_S2}
and Fig.\ \ref{fig:_ExpSetting_S2}. As we shall see that as $T_{0}$
rises across this range, the performance of Separate-RL increases
dramatically and Cohesion-RL\#1 rises gradually, while Cohesion-RL\#2
remains stable. Thus, the average advantage of our method over Separate-RL
decreases dramatically as $T_{0}$ rises, i.e., from $224.07$ steps
and $261.67$ steps when $T_{0}=5$ to $33.26$ steps and $52.09$
steps when $T_{0}=20$. Such case suggests that our methods work perfectly
when the WST is very short. In this case, the mining of network cohesion
is necessary for the online RL learning. In general, however, our
methods still outperform Separate-RL significantly.

\begin{figure*}
\centering{}\includegraphics[width=0.94\linewidth]{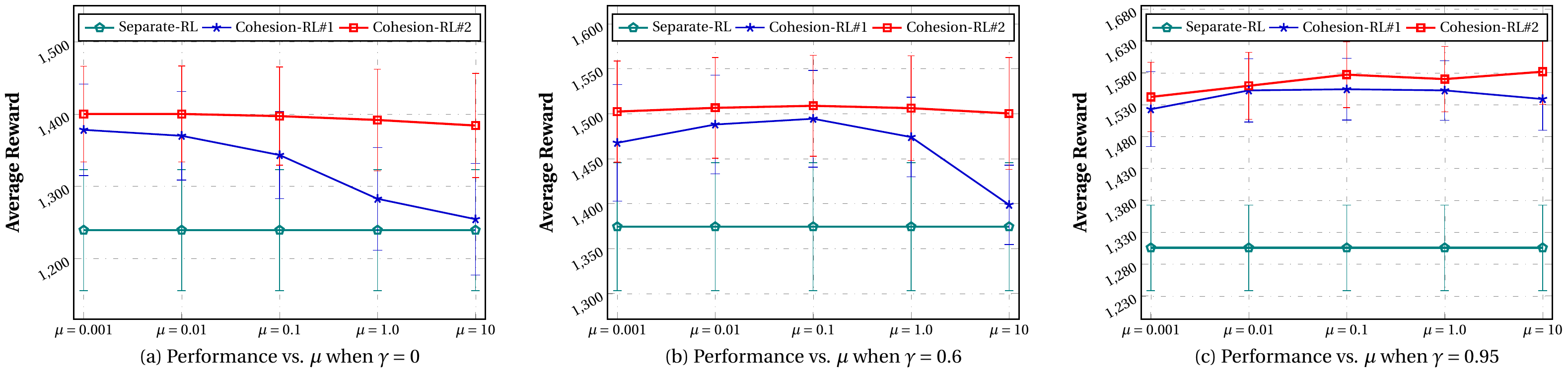}
\caption{Performance of three RL methods for experiment setting (\textbf{S3}).
Each sub-figure shows results under one $\gamma$ setting. \label{fig:_ExpSetting_S3}}
\end{figure*}

(\textbf{S3}) The parameter of the Network-Cohesion constraint $\mu_{1}$
for the projection step ranges from $0.001$ to $10$. To reduce the
number of parameters in our algorithm, we simply set $\mu_{2}=0.01\mu_{1}$
(i.e. the cohesion constraint for the fixed-point step) and $\mu_{3}=\mu_{1}$
(i.e. the cohesion constraint for the actor updating). The experiment
results are illustrated in Fig.\ \ref{fig:_ExpSetting_S3}, where
there are three sub-figures. Each sub-figure shows the results of
three online RLs vs. five $\mu_{1}$ settings under one $\gamma$.
As we shall see that as $\mu_{1}$ rises across this range, our method
always obtains superior performance compared with Separate-RL. Specially,
Cohesion-RL\#2 is very stable and always better than Cohesion-RL\#1.
Such case indicates that it is reliable to follow the idea on how
to introduce the $\ell_{2}$ constraint in LSTD$Q$. In Fig.\ \ref{fig:_ExpSetting_S3},
since Separate-RL does not have the Network-Cohesion constraint, its
result keeps unchanged.

Consider (\textbf{S1}) and (\textbf{S2}) for the Separate-RL, we find:
(a) the lack of samples at the beginning of the online learning may
bias the optimization direction, which badly influence the performance
even when the trajectory is very long; (b) Compared with $T$, the
increase of $T_{0}$ has a more important influence on the performance.
In (\textbf{S1}), where $T_{0}=10$ is fixed and $T$ ranges from
$T=50$ to $T=150$, the performance of Separate-RL increases $32.74$
steps. In (\textbf{S2}), where $T=80$ is fixed and $T_{0}$ rises
from $T_{0}=5$ to $T_{0}=20$, Separate-RL achieve an improvement
of $210.51$ steps, which is much significant than the rise caused
by the rising $T$. 

\section{Conclusions and Discuss \label{sec:Conclusions}}

This paper presents a first attempt to employ the online actor-critic
reinforcement learning for the mHealth. Following the current methods
that learn a separate policy for each user, the Separate-RL can not
achieve satisfactory results. This is due to that data for each user
is very limited in size to support the separate learning, leading
to unstable policies that contain lots of variances. After considering
the universal phenomenon that users are generally connected in a network
and linked users tend to have similar behaviors, we propose a network
cohesion constrained actor-critic reinforcement learning for mHealth.
It is able to share the information among similar users to convert
the limited user information into sharper learned policies. Extensive
experiment results demonstrate that our methods outperform the Separate-RL
significantly. We find it easy to apply the proposed methods to other
health-related tasks.

\subsection*{Appendix: the proof of Theorem\ \ref{thm:B_sysmmetirc_positive_definite} }
\begin{IEEEproof}
Considering $\mtbfL_{\otimes}\left(\mu_{2},\zeta_{2}\right)=\left(\mu_{2}\mtbfL^{\intercal}+\zeta_{2}\mtbfI_{N}\right)\otimes\mtbfI_{u}$
gives the equation $\mtbfB=\mtbfP^{\intercal}\mtbfP+\mu_{2}\mtbfL^{\intercal}\otimes\mtbfI_{u}+\zeta_{2}\mtbfI_{uN}$.
The first term $\mtbfB_{1}=\mtbfP^{\intercal}\mtbfP$ is obviously
positive semi-definite as $\forall\mtbfx$, we have $\mtbfx^{\intercal}\mtbfP^{\intercal}\mtbfP\mtbfx=\left\Vert \mtbfP\mtbfx\right\Vert _{2}^{2}\geq0$.
The graph laplacian $\mtbfL$ is positive semi-definite, which indicates
that its eigenvalues are non-negative, i.e. $\lambda_{1},\cdots,\lambda_{N}\geq0$.
The eigenvalues of $\mtbfI_{u}$ are $\mu_{1}=\cdots=\mu_{u}=1$.
According to Lemma\ \ref{lem:KroneckerProduct_eigenvalues}, we have
the conclusion that the eigenvalues of $\mtbfL^{\intercal}\otimes\mtbfI_{u}$
are non-negative, which indicates that it is a positive semi-definite
matrix. The last term in $\mtbfB$ is an identical matrix, which is
surely positive definite. The sum of two positive semi-definite matrices
and a positive definite matrix results in a positive definite matrix.
\end{IEEEproof}
Since for any matrices $\mtbfA\in\mtbbR^{l\times k}$ and $\mtbfD\in\mtbbR^{m\times n}$,
the Kronecker product has the property $\left(\mtbfA\otimes\mtbfD\right)^{\intercal}=\left(\mtbfA^{\intercal}\otimes\mtbfD^{\intercal}\right)$\ \cite{Langville_2004_JCAM_KroneckerProduct}.
Besides, the graph laplacian $\mtbfL$ is symmetric. We have 
\[
\mtbfB^{\intercal}=\left(\mtbfP^{\intercal}\mtbfP+\mu_{2}\mtbfL^{\intercal}\otimes\mtbfI_{u}+\zeta_{2}\mtbfI_{uN}\right)^{\intercal}=\mtbfB.
\]

\bibliographystyle{ieeetr}
\bibliography{4_home_fyzhu_link2dropbox_self_Folder_myWorksOnDropboxs_bibFiles_referenceBib2,5_home_fyzhu_link2dropbox_self_Folder_myWorksOnDropboxs_bibFiles_referenceBib}

\end{document}